\documentclass[lettersize,journal]{IEEEtran}
\usepackage{graphicx} 
\usepackage{xcolor}
\usepackage{bm} 
\usepackage{amssymb}  
\usepackage{enumitem}
\usepackage{subfigure}
\usepackage{fontenc}
\usepackage{algorithm}
\usepackage{algorithmic}
\newcommand\algorithmicprocedure{\textbf{procedure}}
\newcommand{\algorithmicendprocedure}{\algorithmicend\ \algorithmicprocedure}
\makeatletter
\newcommand\PROCEDURE[3][default]{%
  \ALC@it
  \algorithmicprocedure\ {#2}(#3)%
  \ALC@com{#1}%
  \begin{ALC@prc}%
}
\newcommand\ENDPROCEDURE{%
  \end{ALC@prc}%
  \ifthenelse{\boolean{ALC@noend}}{}{%
    \ALC@it\algorithmicendprocedure
  }%
}
\newenvironment{ALC@prc}{\begin{ALC@g}}{\end{ALC@g}}
\makeatother
\usepackage{cite}
\usepackage{amsmath,amssymb,amsfonts}

\usepackage{graphicx}
\usepackage{textcomp}
\usepackage{xcolor}
\usepackage{mathtools}
\usepackage{amsmath}
\usepackage{amsfonts}
\usepackage{tabularx}
\usepackage{multirow}
\usepackage{multicol}
\usepackage{float}
\usepackage{tabularx}
\usepackage{threeparttable}
\usepackage{booktabs}
\usepackage{longtable}
\usepackage{booktabs}
\usepackage{makecell}
\usepackage{verbatim}
\usepackage{wrapfig}
\usepackage{xspace}

\usepackage{hyperref}

\begin{document}

\title{DexSim2Real$^\textbf{2}$: Building Explicit World Model for Precise Articulated Object Dexterous Manipulation}

\author{Taoran Jiang*, Yixuan Guan*, Liqian Ma*, Jing Xu*~\IEEEmembership{Member,~IEEE,} Jiaojiao Meng, Weihang Chen, Zecui Zeng, Lusong Li, Dan Wu,  Rui Chen~\IEEEmembership{Member,~IEEE}
\thanks{*Taoran Jiang, Yixuan Guan, Liqian Ma, and Jing Xu contributed equally to this work. Corresponding author: Rui Chen chenruithu@mail.tsinghua.edu.cn}
\thanks{Taoran Jiang, Yixuan Guan, Liqian Ma, Jiaojiao Meng, Weihang Chen, Dan Wu, Jing Xu, Rui Chen are with the Department of Mechanical Engineering, Tsinghua University, Beijing 100084, China.}
\thanks{ Liqian Ma is also with the Institute for
Robotics and Intelligent Machines (IRIM), Georgia Institute of Technology. This work was done while he was affiliated with the Department of Mechanical Engineering, Tsinghua University.}
\thanks{Zecui Zeng, Lusong Li are with JD Explore Academy, Beijing, China.}
}



\maketitle

\begin{abstract}

Articulated objects are ubiquitous in daily life. In this paper, we present DexSim2Real$^\textbf{2}$, a novel framework for goal-conditioned articulated object manipulation. 
The core of our framework is constructing an explicit world model of unseen articulated objects through active interactions, which enables sampling-based model predictive control to plan trajectories achieving different goals without requiring demonstrations or RL.
It first predicts an interaction using an affordance network trained on self-supervised interaction data or videos of human manipulation. {After executing the interactions on the real robot to move the object parts, 
we propose a novel modeling pipeline based on 3D AIGC to build a digital twin of the object in simulation from multiple frames of observations.}
For dexterous hands, we utilize eigengrasp to reduce the action dimension, enabling more efficient trajectory searching. Experiments validate the framework's effectiveness for precise manipulation using {a suction gripper, a two-finger gripper and two dexterous hands}. 
The generalizability of the explicit world model also enables advanced manipulation strategies like manipulating with tools.

\end{abstract}


\section{INTRODUCTION}

\IEEEPARstart{A}{rticulated} object manipulation is a fundamental and challenging problem in robotics.
Compared with pick-and-place tasks, where only the start and final poses of robot end effectors are constrained, articulated object manipulation requires the robot end effector to move along certain trajectories, making the problem significantly more complex.
Most existing works utilize a neural network to learn the correlation between object states and correct actions, and employ reinforcement learning (RL) and imitation learning (IL) to train the neural network~\cite{chi2024universal, partrl, wang2022adaafford}. However, since the state distribution of articulated objects is higher-dimensional and more complex than that of rigid objects, it is difficult for the neural network to learn such correlation, even with hundreds of successful demonstrations and millions of interactions~\cite{mu2maniskill, gumaniskill2, tao2024maniskill3}.

\begin{figure}[t]
      \centering
      \includegraphics[width=7.8cm]{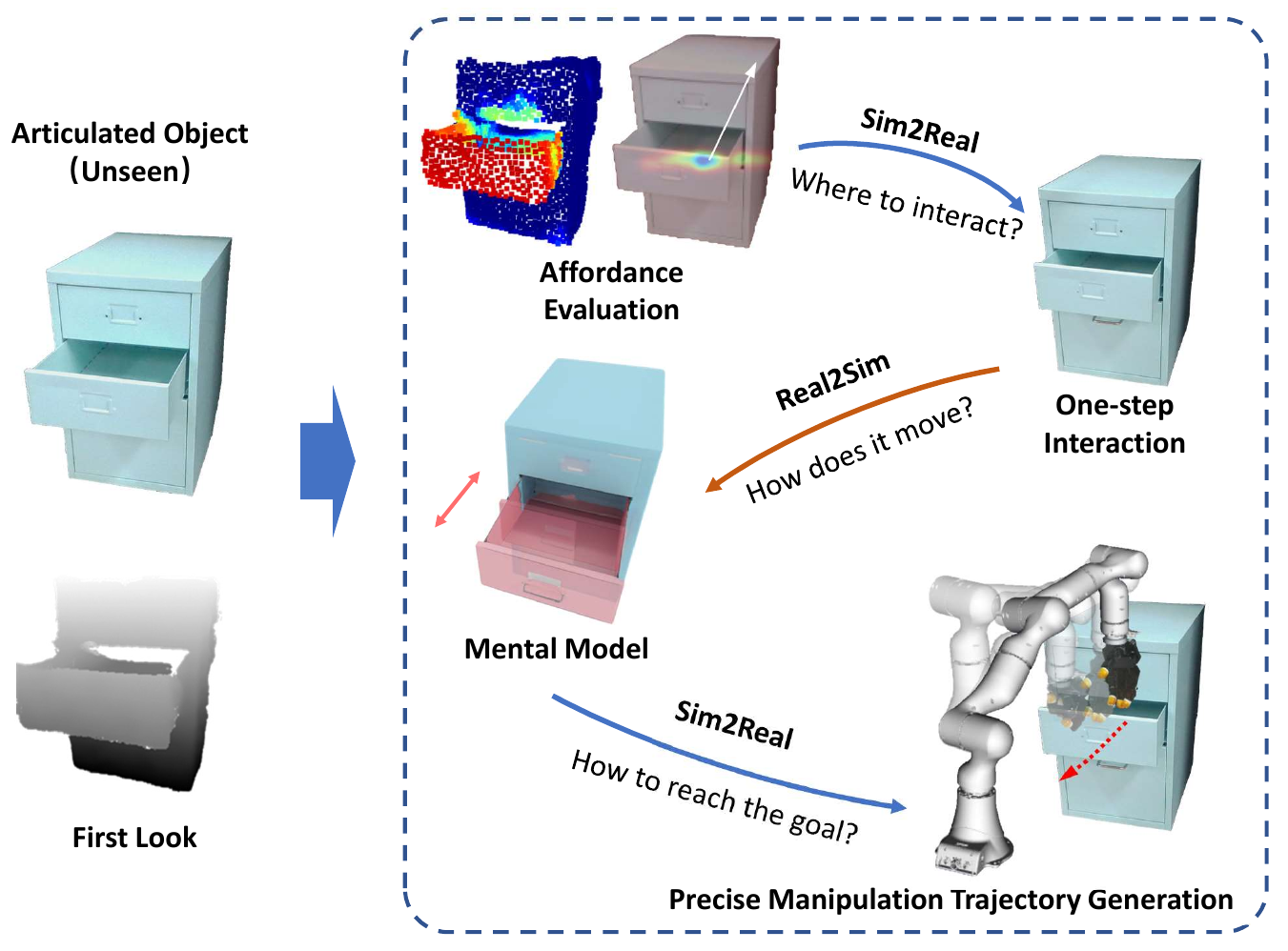}
      \caption{\textbf{DexSim2Real$^\textbf{2}$} is a robot learning framework for precise goal-conditioned articulated object manipulation with suction grippers, two-finger grippers, and multi-finger dexterous hands in the real world. It builds the mental model of the unseen target object through active interactions and uses the model to generate a long-horizon manipulation trajectory.}
      \label{fig:teaser}
\end{figure}
For humans, manipulation involves not only action responding to perception, as is the case with policy networks, but also motor imagery and mental simulation, that humans can imagine the action consequences before execution and plan the action trajectory accordingly\cite{johnson2000thinking}.
To model the world more accurately, humans can actively interact with the environment, change its state and gather additional information, which is named as \textit{interactive perception}~\cite{von2007action, hofsten2009action}.

In this paper, we propose a robot learning framework called DexSim2Real$^\textbf{2}$  to achieve precise goal-conditioned manipulation of articulated objects using  {multiple types of end effectors, including suction grippers, two-finger grippers, and dexterous hands}, where we use a physics simulator as the \textit{mental model} of robots. 
Fig.~\ref{fig:teaser} provides a brief overview of our framework. The framework first learns an affordance estimation network from self-supervised interaction in simulation or egocentric videos of human-object interactions.  The network predicts a one-step motion of the robot end effector from a single-frame observation of the object. The reason why we first learn the affordance is that affordance estimation is only attributed to the object and can better generalize to novel objects. Also the one-step interaction does not require fine manipulation of the end effector. Next, we execute the predicted action on the real robot to change the object's state and capture another observation after the interaction.
{For an articulated object with $K$ movable parts, this step is repeated by $K$ times such that the positions of all the movable parts are changed.}
{Then, we propose a novel pipeline based on 3D AIGC (Artificial Intelligence Generated Content) and foundation vision models to construct an explicit world model of the articulated object including the shapes of all the parts and the kinematic structure from the $K+1$ observations. }
Finally, using the explicit world model we have built, we utilize sampling-based model predictive control (MPC) to plan a trajectory to achieve goal-conditioned manipulation tasks.

While dexterous manipulation with multi-finger hands enables more flexible, efficient and robust manipulation, the high-dimensional action space presents significant challenges for MPC. To handle this problem, we propose to employ eigengrasp~\cite{ciocarlie2007dexterous} to reduce the operational dimensions of the dexterous hand, enabling more efficient and successful searching.  While eigengrasp has been widely studied for robot grasping~\cite{9256325, 10225433, agarwal2023dexterous}, its application in dexterous manipulation remains under-explored. 
By leveraging eigengrasp and the constructed explicit world model, we can accurately predict the motion of the articulated object upon contact with the dexterous hand and search for a feasible dexterous manipulation trajectory.

This article is an extension of our previous ICRA work: Sim2Real$^\textbf{2}$~\cite{10160370}. There are {three} main additional contributions in this work:

{(1) We propose a novel articulated object model construction pipeline. Compared with Ditto~\cite{jiang2022ditto}, our pipeline is able to handle objects with more than one movable parts. Moreover, since we utilize 3D AIGC to reconstruct the geometry, the reconstruction quality is significantly improved. }

(2) {We broaden the framework's scope from manipulation with two-finger gripper to more types of end effectors, including suction gripper and multi-finger dexterous hand.} To address the challenge introduced by the high-dimensional action space of the dexterous hand, we propose to utilize eigengrasp to reduce the dimension, leading to more efficient and successful manipulation. We conduct extensive experiments both in simulation and on a real robot to validate our method's effectiveness for dexterous manipulation and the usefulness of its different modules.

(3) In our previous work, we use self-supervised interaction in simulation to generate training data for affordance estimation, which requires interactable 3D assets which is still inadequate currently. To eliminate such dependency and enhance our framework's scalability, we propose to learn the affordance from egocentric human manipulation videos, which are large-scale and freely accessible. However, since trajectories in videos are in 2D pixel space, we propose a spatial projection method to generate 3D robot motions from 2D trajectory predictions.

The remainder of this paper is structured as follows: Related works are reviewed in Section \ref{sec:related_work}. Our proposed robot learning framework is detailed in Section \ref{sec:method}. Experimental setup and results are presented in Section \ref{sec:experiments}. {Experiment analysis and limitations are discussed in Section~\ref{sec:analysis}.} Finally, our work and future directions are concluded in Section \ref{sec:conclusion}.

\section{RELATED WORK}
\label{sec:related_work}

{
\subsection{Learning-Based Articulated Object Manipulation}
Articulated object manipulation remains an open and challenging research problem due to the complex kinematic constraint. With the development to 3D learning algorithms, larger-scale articulated object dataset and robot physics simulation platform, many learning-based works have been proposed for affordance learning~\cite{mo2021where2act, ning2024where2explore}, articulation parameter estimation~\cite{wang2019shape2motion, jain2021screwnet, nie2023structure}, and manipulation skill learning~\cite{gumaniskill2,EisnerZhang2022FLOW,xu2021umpnet, zhang2023flowbot, partrl}. 

In FlowBot~\cite{EisnerZhang2022FLOW}, the per-point 3D articulation flow of the point cloud is first estimated, then the point with the maximum flow vector magnitude is chosen for manipulation with a suction gripper. FlowBot++~\cite{zhang2023flowbot} enhances this process by explicitly estimating the articulation parameters, which improves the smoothness of manipulation. In UMPNet~\cite{xu2021umpnet}, the action is predicted based on two RGBD images: one of the current state and the other of the initial or goal state. Therefore, UMPNet is suitable for goal-conditioned manipulation, aligning closely with the task addressed in our work.

Our approach, however, distinguishes itself from the above methods in three key aspects: (1) While these works rely on single-frame observations to predict actions, our method leverages multi-frame observations through active interaction with the object, enabling more accurate articulation structure estimation. (2) These works use a suction gripper to establish a stable contact with the object and cannot be easily extended to other end-effectors. In contrast, our method supports multiple kinds of end effectors, including suction grippers, two-finger grippers and multi-finger hands. (3) In contrast to articulation flow which represents a one-step object motion in current state, our method builds an explicit model of the articulated object within a physics simulation, which can generate long-horizon multi-step manipulation trajectories prior to physical interaction, and further generalize to unseen interactions, like manipulation with tools demonstrated in our experiments.
}

\subsection{Dexterous Manipulation}
Compared with two-finger grippers, multi-finger dexterous hands can manipulate a broader range of objects with more human-like dexterous actions~\cite{844067}. Traditional model-based approaches formulate dexterous manipulation as a planning problem and generate trajectories through search and optimization~\cite{rus1999hand, dogar2010push, kumar2016optimal, wu2022learning}. 
These methods require accurate 3D shapes of the manipulated object and the hand, which limits their applicability to unseen objects.

In contrast, data-driven methods learn manipulation policies through imitation learning and reinforcement learning~\cite{gupta2016learning, radosavovic2021state, 9849105, andrychowicz2020learning, pmlr-v205-qin23a,  chen2022system}. In~\cite{9849105}, a single-camera teleoperation system is developed for 3D demonstration trajectory collection, significantly reducing the equipment cost. Nevertheless, the time consuming nature of human demonstration and the space required for scene setup still limits the scalability of imitation learning. RL eliminates the need for demonstrations and leads to better scalability. Most existing RL methods learn a policy, which directly maps the observation into the joint angles of the dexterous hand~\cite{pmlr-v205-qin23a, chen2022system, andrychowicz2020learning}. However, the high-dimensional action space slows the learning efficiency and usually results in uncommon hand motion which cannot be executed on real robot hands. In~\cite{agarwal2023dexterous}, eigengrasps~\cite{ciocarlie2007dexterous} are used to reduce the dimension of the action space for functional grasping. Experimental results show that the utilization of eigengrasp can lead to more stable and physically realistic hand motion for robot grasping. 
However, more advanced manipulation policies are not studied in this work. 

In our work, we combine the advantages of model-based methods and data-driven methods by first learning a generalizable world model construction module and then using the model to search for a feasible trajectory for dexterous manipulation. Furthermore, we adopt eigengrasps to accelerate the searching process and generate more reasonable hand motions that can be directly executed on real robots.

\subsection{World Model Construction}

Building an accurate and generalizable transition model of the environment capable of reacting to agent interactions has been a long-standing problem in optimal control and model-based RL~\cite{branicky1998unified, sutton1991dyna}. Some existing methods model the dynamic system in a lower-dimensional state space, reducing computation and simplifying the transition model~\cite{achille2018separation, castro2020scalable, hansen2022temporal}. However, this approach discards the environment's spatial structure, which limits the model's generalizablity to novel interactions. 

With increasing computational power and large network architectures, image-based and video-based world models have gain increasing attention~\cite{finn2016unsupervised, finn2017deep, wu2023daydreamer, yanglearning}. In~\cite{yanglearning}, a U-Net-based video diffusion model is used to predict future observation video sequence from the past observations and actions. While it shows great ability to emulate real-world manipulation and navigation environments, it requires an extremely large-scale dataset and computational resources for network training, because the network contains minimal knowledge prior of the environment. Additionally, the inference speed of the large network limits its feasibility for MPC. 

In our work, we focus on articulated object manipulation, so we introduce the knowledge prior of the environment by using an explicit physics model.
Therefore, we are able to significantly decrease the number of samples required for model construction.
Moreover, the explicit physics model's generalizability guarantees that while we only use a simple action to collect the sample, the built model can be used for long-horizon complex trajectory planning composed of unseen robot actions.

{
Articulated object modeling has recently gained significant attention in the fields of robotics and computer vision~\cite{jiang2022ditto,liu2024survey,mandi2024real2code,weng2024neural}. Ditto~\cite{jiang2022ditto} employs a 3D network to predict the voxel occupancy and joint parameters simultaneously using two frames of point clouds as input. However, Ditto is limited to objects with a single joint. Liu et al.~\cite{liu2023building} propose to build kinematic structures from point cloud sequences by jointly optimizing part segmentation, transformation, and joint parameters through an energy minimization framework. But the performance of this approach heavily depends on part segmentation quality, which is not always reliable when the sequence only consists of 2-3 frames.
In our work, we address this limitation by designing a novel movable part segmentation module. Our module leverages mesh connectivity, foundation vision models, and proprioception from robotic interactions to achieve more accurate and robust segmentation, leading to better kinematic structure estimation.
More recently, Articulate-Anything~\cite{le2024articulate} explore the use of vision-language models to generate programs for constructing Unified Robot Description Format (URDF) files from text, images, and videos. However, their approach relies on mesh retrieval from pre-existing datasets to define object parts, which limits geometric diversity and alignment with real-world observations. In contrast, we leverage advanced 3D AIGC techniques to reconstruct meshes that align precisely with observations. This enables accurate estimation of pose, position, and scale, further supporting precise manipulation in our proposed Sim2Real${}^2$ pipeline.
}

\subsection{Affordance Learning}
In the context of articulated objects, affordances dictate how their movable parts can be interacted by a robot to achieve a desired configuration, which provides a valuable guide for articulated object manipulation. Therefore, affordance learning has been widely studied in the literature. Deng et al. built a benchmark for visual object affordance understanding by manual annotation~\cite{Deng_2021_CVPR}. Cui et al. explored learning affordances using point supervision~\cite{cui2023strap}.  While these supervised learning methods can yield accurate affordance predictions, the cost of the manual annotation process limits their scalability. 

Another line of research focuses on learning the affordances through interactions in simulation~\cite{mo2021where2act, 9681198, wang2022adaafford}. Where2Act~\cite{mo2021where2act} first collects random offline interaction trajectories and then samples online interaction data points for training data generation to facilitate affordance learning. However, the key bottleneck of simulation-based methods is the requirement for 3D articulated object assets that can be accurately interacted with and simulated. Unfortunately, most existing 3D object datasets only include static CAD models, which cannot be used for physics simulation~\cite{deitke2023objaverse, deitke2024objaverse}.

Videos of human-object interactions are free, large-scale, and diverse, making them an ideal data source for robot learning~\cite{nagarajan2019grounded,Bahl_2023_CVPR , bharadhwaj2024track2act}. In VRB~\cite{Bahl_2023_CVPR}, the contact point and post-contact trajectory are first extracted from videos of human manipulation, and then they are used to supervise the training of the affordance model. However, the predicted affordance is only 2D coordinate and direction in the image, which cannot be directly used for robot execution. Therefore, we propose to generate the robot interaction direction in the 3D physical space by synthesizing a virtual image from the RGBD data and computing the 3D robot motion as the intersection of the 2 VRB predictions in the 3D space.
\begin{figure*}[thpb]
    \centering
    \includegraphics[width=17cm]{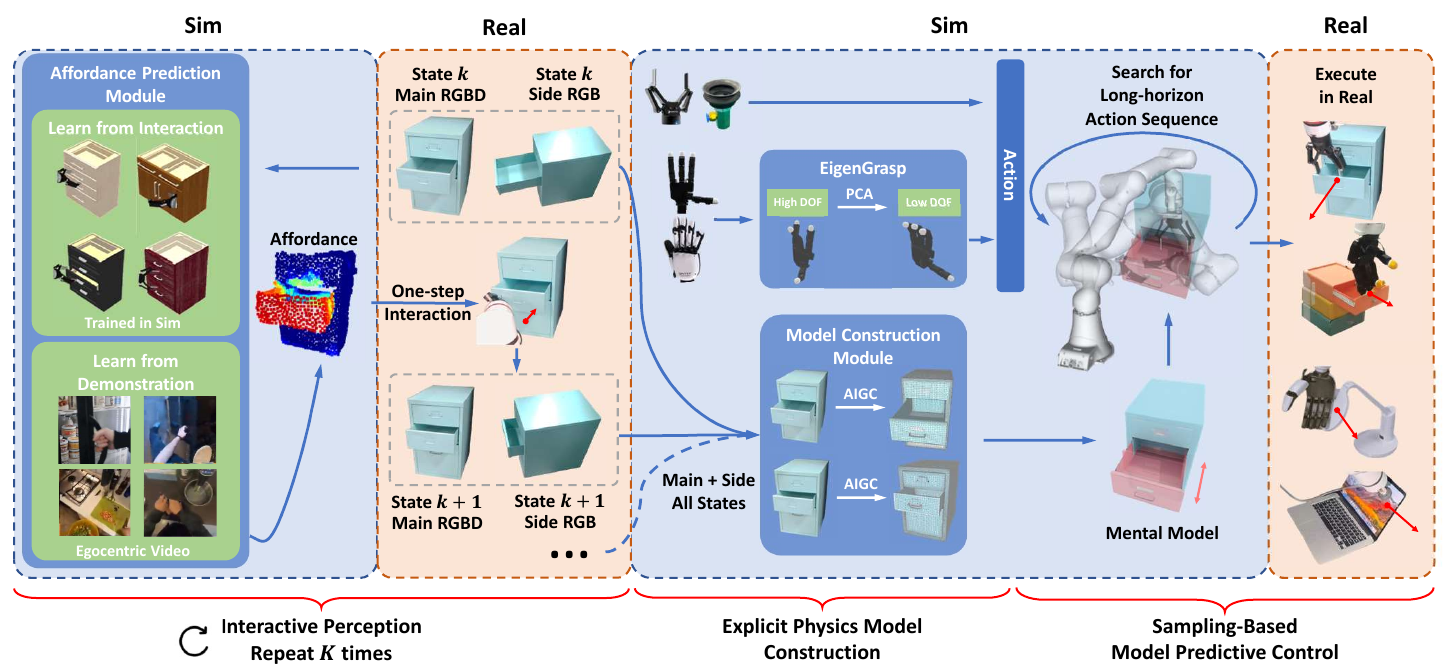}
    \caption{Overview of the \textbf{DexSim2Real$^\textbf{2}$} framework. Our framework consists of three phases. (1) Given a partial point cloud of an unseen articulated object, in the \textbf{Interactive Perception} phase, we train an affordance prediction module and use it to change the object’s joint state through a one-step interaction. Training data can be acquired through self-supervised interaction in simulation or from egocentric human demonstration videos. (2) In the \textbf{Explicit Physics Model Construction} phase, we build a mental model in a physics simulator {from the $K+1$ frames of observations}. (3) In the \textbf{Sampling-based Model Predictive Control} phase, we use the model to plan a long-horizon trajectory in simulation and then execute the trajectory on the real robot to complete the task. For dexterous hands, an eigengrasp module is needed for dimensionality reduction.}
    \label{fig:overview}
\end{figure*}
\subsection{Sim2Real for Robot Learning}

Physics simulation plays a pivotal role in manipulation policy learning, offering large-scale parallelism, reduced training costs, and avoidance of potential damage to robots and researchers~\cite{james2020rlbench, zhu2020robosuite, makoviychuk2021isaac, 10388459}. Most existing methods utilize RL for policy learning in simulation and then deploy the learned policy on a real robot~\cite{sadeghi2018sim2real,hofer2021sim2real, dimitropoulos2022brief}. 
DexPoint~\cite{pmlr-v205-qin23a} utilizes the concatenation of observed point clouds and imagined hand point cloud as inputs and learns dexterous manipulation policy. However, since the neural network does not contain any prior knowledge of the environment, a large amount of interaction data is required to improve its accuracy and generalizablity. In contrast, we propose to first build the explicit world model of the target object and employ MPC to generate manipulation trajectories based on the built model of the single object instance. By avoiding the diversity of objects, we substantially reduce the required interactions and improve the manipulation accuracy in the real world.

\section{METHOD}
\label{sec:method}

The goal of our work is goal-conditioned articulated object manipulation, that is to manipulate articulated objects to specified joint states with various robot-effectors in the real world, including {suction grippers,} two-finger grippers, and multi-finger dexterous hands.
Fig.~\ref{fig:overview} shows an overview of our framework. It consists of three modules: \textbf{Interactive Perception}~(Section~\ref{Interactive Perception}), \textbf{Explicit World Model Construction}~(Section~\ref{Explicit World Model Construction}), \textbf{Sampling-Based Model Predictive Control}~(Section~\ref{Sampling Based Model Predictive Control}).

A single observation of an articulated object cannot provide enough information to reveal its full structure. For example, when humans first look at a kitchen door, it is hard to tell whether it has a rotating hinge or a sliding hinge. However, after the door is moved, humans can use the information from the two observations to infer the type and location of the hinge. Inspired by this, the \textbf{Interactive Perception} module proposes an action to alter the joint state of the articulated object based on learned affordances. This action is then executed on the object in the real world, resulting in a changed state and additional frames of observation.

With the multiple frames of observations, the \textbf{Explicit World Model Construction} module (Section~\ref{Explicit World Model Construction}) infers the shape and the kinematic structure of the articulated object to construct a digital model. The digital model can be loaded into a physics simulator for the robot to interact with, forming an explicit world model of the environment.

The constructed world model can be used to search for a trajectory of control inputs that change the state of the articulated object from \( s_{\text{initial}} \) to a target state \( s_{\text{target}} \) using \textbf{Sampling-Based Model Predictive Control}, introduced in Section \ref{Sampling Based Model Predictive Control}. With the model of a specific object, we can efficiently plan a trajectory using sampling-based MPC to manipulate the object precisely, rather than learning a generalizable policy.

\subsection{Interactive Perception}
\label{Interactive Perception}
At the beginning, the articulated object is placed statically within the scene, and the robot has only a single-frame observation of it. Understanding the articulation structure and surface geometry of each part of the object from this limited view is challenging. However, by actively interacting with the object and altering its state, additional information can be gathered to enhance the understanding of its structure.
It is worth noting that the interaction in this step does not require precision. 

To achieve this goal, it is essential to learn to predict the affordance based on the initial single-frame observation.
In our work, we first learn the affordance through self-supervised interaction in simulation. However, simulation requires interactable 3D assets, which are still relatively scarce. Therefore, we further study learning affordances from real-world human manipulation videos, which are readily available and large-scale.

\subsubsection{Learn from self-supervised interaction in simulation}

By extensively interacting with articulated objects in the simulation, actions that change the state of the articulated object to some extent can be automatically labeled as successful. Using these automatically labeled observation-action pairs, neural networks can be trained to predict candidate actions that can change the object's state based on the initial observation of the object. 

For affordance learning in this method, we use Where2Act~\cite{mo2021where2act}. This algorithm includes an Actionability Scoring Module, which predicts an actionability score \(a_p\) for all points. A higher \(a_p\) indicates a higher likelihood that an action executed at that point will move the part. Additionally, the Action Proposal Module suggests actions for a specific point. The Action Scoring Module then predicts the success likelihood of these proposed actions.

In Where2Act, only a flying gripper is considered, and primitive actions are parameterized by the gripper pose in \(SE(3)\) space. This approach does not account for the robot's kinematic structure, increasing the difficulty of execution in the real world due to potential motion planning failures. Although this simplification eases the learning process, it complicates real-world execution, as motion planning may not find feasible solutions for the proposed actions.

To address this problem, we select \(n_p\) points with the highest actionability scores as candidate points. For each candidate point, we choose \(n_a\) actions with the highest success likelihood scores from the proposed actions. We then use motion planning to attempt to generate joint trajectories for these actions sequentially until a successful one is found. Empirically, we find that this method improves the success rate for the motion planner because the action with the highest success likelihood is often outside the robot's dexterous workspace.

\subsubsection{Learn from real-world egocentric demonstrations} 
\label{sec:method_vrb}
Acquiring 3D affordance representations through self-supervised interactions in simulation has shown promise as it does not rely on labeled data. However, certain limitation exists: the success of this method hinges on interactive models in simulation. Unfortunately, the availability of simulated datasets for articulated objects is limited, hindering the generation of training data.

To address this limitation, we propose another approach that leverages real-world egocentric videos of humans interacting with objects. This complementary data source allows us to overcome the limitations of simulation-based learning and broaden the scope of our affordance representation system. Specifically, we utilize the Vision-Robotics Bridge (VRB)~\cite{Bahl_2023_CVPR} to predict the affordance of articulated objects. VRB introduces an innovative affordance model that learns from human videos. It extracts the contact region and post-contact wrist trajectory of the video. These cues serve as supervision signals for training the affordance model. Given an RGB image of an object as input, the VRB model generates two key outputs: a contact heatmap, highlighting the regions where contact occurs, and a 2D vector representation of the post-contact trajectory within the image. Both of these two outputs are within 2D space. However, for effective interaction between robots and objects in the real world, a 3D manipulation strategy is necessary. To address this issue, we need to convert the 2D affordance generated by the model into valid 3D spatial vector and contact region.

\begin{figure}[t]
      \centering
      \subfigure[]
      {
      \includegraphics[width=9cm]{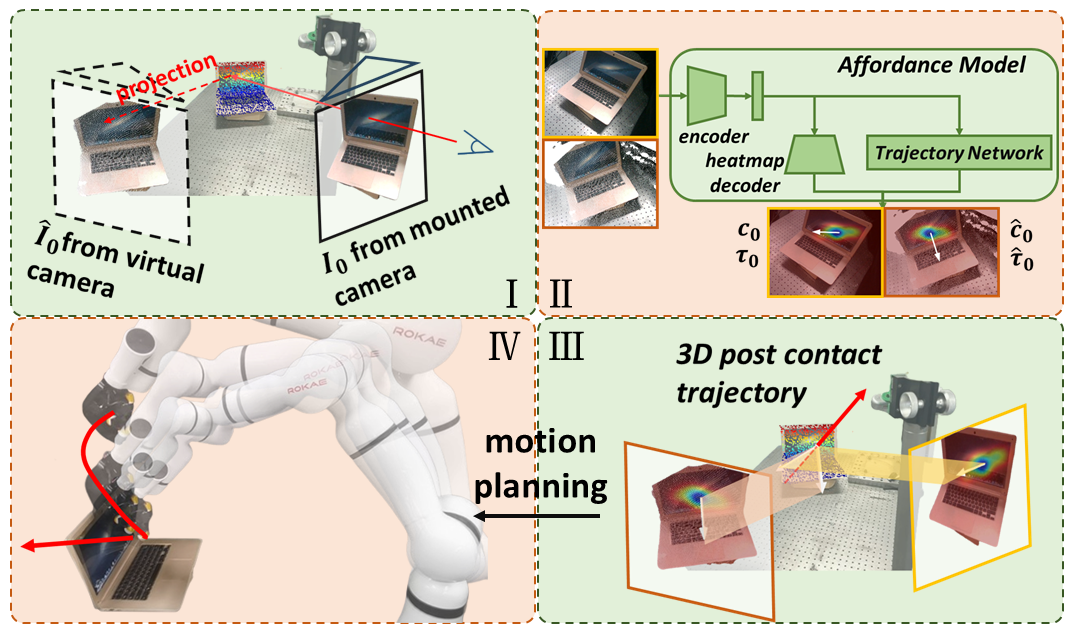}
      \label{fig:vrb_framework}
      }
      \hspace{4mm}
      \subfigure[]
      {
      \includegraphics[width=9cm]{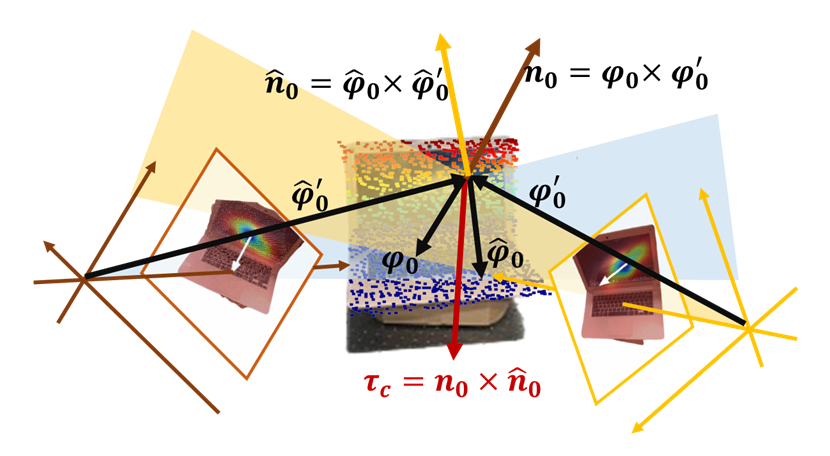}
      \label{fig:vector_calculate}
      }
      \caption{(a)Framework of generating real robot manipulation trajectory from 2D affordances. (b)Calculation method of 3D post contact vector generation.}
      \label{fig:vrb_method}
      
\end{figure}

Fig.~\ref{fig:vrb_framework} illustrates how we generate a 3D trajectory for real robot manipulation from 2D affordances. Firstly, we capture an RGB image $\boldsymbol{I}_0$ and a depth map $\bm{\mathcal{D}}_0$ using the mounted RGBD camera, $\boldsymbol{H}_0\in \mathbb{R}^{4\times4}$ is its relative transformation matrix respect to the robot coordinate system. Secondly, we set a virtual camera, the relative transformation matrix of which is $\boldsymbol{H}\in \mathbb{R}^{4\times4}$. Since the depth of each pixel in $\boldsymbol{I}_0$ is known, we can generate the virtual RGB image $ \hat{\boldsymbol{I}}_0$ by image wrapping. 
Thirdly, we use $\boldsymbol{I}_0$ and $\hat{\boldsymbol{I}}_0$ as the input of the affordance model and generate contact points $\boldsymbol{c}_0=(u_0, v_0)$, $\hat{\boldsymbol{c}}_0=(\hat{u}_0, \hat{v}_0)$ and post-contact trajectories $\boldsymbol{\tau}_0=(u_0^{'}-u_0,v_0^{'}-v_0)$ and $\hat{\boldsymbol{\tau}}_0=(\hat{u}_0^{'}-\hat{u}_0,\hat{v}_0^{'}-\hat{v}_0)$. The camera intrinsic matrix is $\boldsymbol{K}$, the contact point in the mounted camera frame is $\boldsymbol{p}_c \in \mathbb{R}^3$, the 3D post contact vector in the camera frame is $\boldsymbol{\tau}_c \in \mathbb{R}^3$.
Fourthly, we respectively calculate the 3D contact point and post contact vector. We use contact point $\boldsymbol{c}_0$ to acquire 3D contact point $\boldsymbol{p}_c \in \mathbb{R}^3$ in the robot base frame:
\begin{equation}
    \boldsymbol{p}_c
    = \boldsymbol{H}_0^{-1}
      \boldsymbol{K}^{-1}
      {z}_c
    \begin{bmatrix}
        u_{0}\\
        v_{0}\\
        1
    \end{bmatrix}
\end{equation}
where $z_c$ represents the depth of $\boldsymbol{p}_c$.
We use the camera's intrinsic matrix to transfer $\boldsymbol{c}_0$ to point in mounted camera frame {then} use the mounted camera's extrinsic matrix to transfer it to the 3D point cloud $\boldsymbol{p}_c$.

However, generating 3D post-contact vector from 2D information can be comparatively difficult. we can regard the 2D post contact vectors as the projection of 3D vector on their image planes. For each 2D vetcor, there exists countless 3D vectors whose projection on the image plane is the same as the 2D vector. These vectors are all distributed in a same ``projection plane". Given that two different 2D vectors have been generated, we can use the intersection lines of two planes to represent the 3D post contact vector. 

Specifically, our method of calculating 3D post contact vector is shown in Fig.~\ref{fig:vector_calculate}. We respectively denote the projection plane of $\boldsymbol{I}_0$ and $\hat{\boldsymbol{I}}_0$ as $\boldsymbol{S}_0$ and $\hat{\boldsymbol{S}}_0$. For $\boldsymbol{S}_0$, we use $\boldsymbol{\varphi}_0$ and $\boldsymbol{\varphi}_0^{'}$ to represent the projection plane. $\boldsymbol{\varphi}_0$ represents one possible 3D vector on projection plane  $\boldsymbol{S}_0$.  Its starting point is $\boldsymbol{p}_c$, while its ending point can be calculated with:
\begin{equation}
    \boldsymbol{p}_c^{'}
    = \boldsymbol{H}_0^{-1}
      \boldsymbol{K}^{-1}
      {z}_c
    \begin{bmatrix}
        u_0^{'}\\
        v_0^{'}\\
        1
    \end{bmatrix}
\end{equation}
It is worth noticing that within the camera frame, $\boldsymbol{p}_c^{'}$ and $\boldsymbol{p}_c$ share the same depth. $\boldsymbol{\varphi}_0^{'}$ starts from the origin of the camera frame and ends at $\boldsymbol{p}_c$:
\begin{equation}
    \boldsymbol{\varphi}_0
    =\boldsymbol{p}_c^{'}-\boldsymbol{p}_c
\end{equation}
\begin{equation}
    \boldsymbol{\varphi}_0^{'}
    = 
    \boldsymbol{p}_c-\boldsymbol{o}_{c0}
\end{equation}
where $\boldsymbol{o}_{c0}$ is the coordinate of camera frame's origin in the robot base frame.
Then we calculate the norm vector of $\boldsymbol{S}_0$: $\boldsymbol{n}_0$ = $\boldsymbol{\varphi}_0\times\boldsymbol{\varphi}_0^{'}$. We can calculate $\hat{\boldsymbol{n}}_0$ in the same way. Finally, we generate the 3D post-contact vector in the robot base frame: $\boldsymbol{\tau}_c$ = $\boldsymbol{n}_0\times\hat{\boldsymbol{n}}_0$.

Finally, we use motion planning to conduct the one-step interaction with the articulated object. The motion planning process can be divided into two phases: we first moves the end effector to the contact point and then we control the end effector to move a certain distance in the direction of the post contact vector. {Since the purpose of the one-step interaction is to change the state of the movable part, a pushing interaction is sufficient in most cases. For objects where pushing is not feasible—such as fully closed cabinets and drawers—
we take an alternative grasping approach. We first utilize VRB to detect the handle. Next, we open the effector and move it to a position near, but behind, the handle. Then, we move the effector to the handle and close it to grasp the handle. Finally, we move a certain distance in the direction of the post-contact vector.
For objects with $K > 1$ movable parts, we iteratively detect affordance points by identifying one affordance point, masking out its neighborhood, and then detecting the next affordance point. This process is repeated until $K$ affordance points are identified.
}

{
\subsection{Explicit World Model Construction}
\label{Explicit World Model Construction}
\begin{figure*}
    \centering
    \includegraphics[width=17cm]{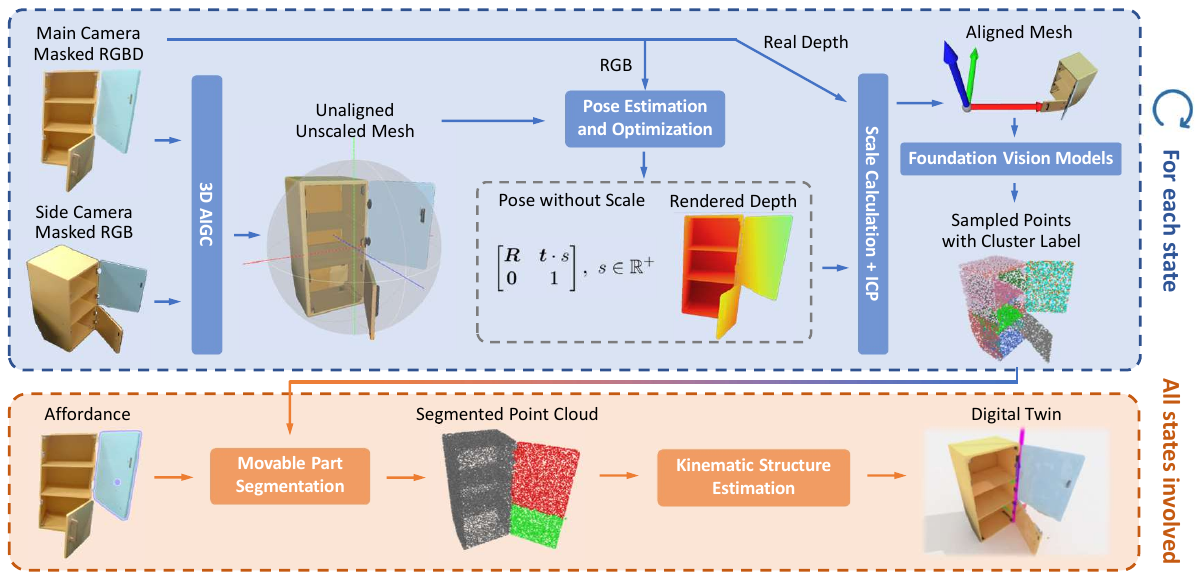}
    \caption{{
    Our pipeline for explicit world model construction: 
    For each state of the articulated object, we begin by generating an unaligned and unscaled mesh from multi-view RGB images using 3D AIGC. Next, we estimate the scale and pose through differentiable rendering, and segment the aligned mesh into sub-parts. Once segmented point clouds for each state are obtained, we infer movable part segmentation by analyzing differences between frames of point clouds. We estimate the kinematic structure of the mesh, including the part tree hierarchy, joint categories (prismatic or revolute), and joint configurations (axis direction and origin). Finally, we construct a digital twin of the articulated object represented in URDF format, which can be easily loaded into different physics simulators.
    }}
    \label{fig:model_construction}
\end{figure*}
In this section, we propose a novel explicit world model construction method that we reconstruct the geometry and the kinematic structure of an articulated object.
Fig.~\ref{fig:model_construction} shows the whole pipeline. 
We assume the object consists of one root part and $K$ movable parts. 
The input are $K+1$ frames of observations:\{$\boldsymbol{I}^{(k)}_0$, $\boldsymbol{D}^{(k)}_0$, $\boldsymbol{M}^{(k)}_0$, $\boldsymbol{I}^{(k)}_1$,  $\boldsymbol{M}^{(k)}_1$,..., $\boldsymbol{I}^{(k)}_C$,  $\boldsymbol{M}^{(k)}_C$\}, $k=0,...,K$. 
$\boldsymbol{I}^{(k)}_0$, $\boldsymbol{D}^{(k)}_0$ are the RGB image and depth map from the main RGBD sensor, and $\boldsymbol{M}^{(k)}_0$ is the object segmentation mask of $\boldsymbol{I}^{(k)}_0$. Besides the main sensor, we also use $C$ RGB cameras located at different viewpoints to improve the reconstruction quality. $\boldsymbol{I}^{(k)}_c$ is the RGB image captured by the $c$th camera, and $\boldsymbol{M}^{(k)}_c$ is the corresponding object segmentation mask. In this work, we use $C=1$.

\subsubsection{Canonical Mesh Generation}
The first step of the pipeline is to reconstruct the $k$th frame canonical mesh $\bm{\widetilde{\mathcal{A}}}^{(k)}$. Here we utilize a controllable large-scale generative model, Rodin~\cite{hyper3d}, to generate $\bm{\widetilde{\mathcal{A}}}^{(k)}$ conditioned on \{$\boldsymbol{I}^{(k)}_0$, $\boldsymbol{M}^{(k)}_0$, $\boldsymbol{I}^{(k)}_1$,  $\boldsymbol{M}^{(k)}_1$,..., $\boldsymbol{I}^{(k)}_C$,  $\boldsymbol{M}^{(k)}_C$\}.

\subsubsection{Mesh Positioning and Scaling}
While the generated mesh $\bm{\widetilde{\mathcal{A}}}^{(k)}$ exhibits high-quality geometry and texture, it is positioned at the origin and scaled to a unit size. Therefore, it is required to obtain the real scale and position of the object for further operation.

To achieve this, we first estimate the position of the object. Since $\bm{\widetilde{\mathcal{A}}}^{(k)}$ is scaled to unit size, we cannot directly apply CAD-model-based 6D pose estimation methods like FoundationPose~\cite{wen2024foundationpose}. Therefore, we modify FoundationPose to meet our specific requirements. Specifically, we first render a set of images of $\bm{\widetilde{\mathcal{A}}}^{(k)}$ with different pose hypotheses and sort these poses based on the images' similarity to $\boldsymbol{I}^{(k)}_0$. Secondly, for the pose with the highest similarity, we refine it by aligning the rendered edges with the edges of $\boldsymbol{M}^{(k)}_0$ using differentiable rendering~\cite{laine2020modular}. To improve the stability of the optimization, we constrain the orientation to the $Z$-axis. 

After determining the orientation $\bm R$ and the unscaled translation $\bm {\widetilde{t}}$ , we can render an unscaled depth map $\widetilde{\boldsymbol{D}}^{(k)}_0$. Then the scale $s^{(k)}$ can be computed as:

\begin{equation}
    s^{(k)} = \frac{\sum_{\boldsymbol{M}^{(k)}_0}\boldsymbol{D}^{(k)}_0}{\sum_{\boldsymbol{M}^{(k)}_0}\widetilde{\boldsymbol{D}}^{(k)}_0}
\end{equation}
Once $s^{(k)}$ is computed, we can generate a correctly scaled mesh $\bm{{\mathcal{A}}}^{(k)}$ and position it accurately to align with its real-world location.

\subsubsection{Movable Part Segmentation}
To build an interactable model of the object, we need to segment the mesh $\bm{{\mathcal{A}}}^{(k)}$ into movable parts.
However, it is difficult to distinguish the parts from a single-frame static mesh. Therefore, we use all the $K+1$ meshes with different kinematic states to segment the parts.  The output is a point cloud $\bm{{\mathcal{P}}}^{(K)}\in \mathbb{R}^{N^{(K)}\times3}$ sampled from $\bm{{\mathcal{A}}}^{(K)}$ with part segmentation labels $\bm{{\mathcal{S}}}^{(K)}\in \{0,1,...,K\}^{N^{(K)}}$, where $N^{(K)}$ is the number of points of $\bm{{\mathcal{P}}}^{(K)}$.

We first segment the mesh into small sub-parts $\bm{{\mathcal{Q}}}^{(k)}$ using foundation vision models. In this work, we choose SAMPart3D~\cite{yang2024sampart3d}. To determine whether each sub-part is moved, we find the sub-part $\bm{P}$ 's closest point set $\bm{PB}$ in $\bm{{\mathcal{P}}}^{(k-1)}$  and compute the chamfer distance $\{d_i^{(k)\rightarrow(k-1)}\}$, then we find $\bm{PB}$'s closest points $\bm{PF}$ in $\bm{{\mathcal{P}}}^{(k)}$ and compute the distances $\{d_i^{(k-1)\rightarrow(k)}\}$.  We set $\bm{P}$ as moved if $\{d_i^{(k)\rightarrow(k-1)}\}$ is larger than certain threshold or the distribution of $\{d_i^{(k)\rightarrow(k-1)}\}$ and $\{d_i^{(k-1)\rightarrow(k)}\}$ are different. However, this criteria may result in wrong segmentation due to the inconsistency of 3D AIGC for different frames or the error of pose estimation. 

To further improve the robustness, we utilize the sub-part connectivity from the mesh and the robot proprioception. For sub-parts $\bm{{\mathcal{Q}}}^{(k)}$ of $\bm{{\mathcal{A}}}^{(k)}$, we can compute the connectivity between each sub-part pair and build a graph $\bm{{\mathcal{G}}}^{(k)} = <\bm{{\mathcal{Q}}}^{(k)}, \bm{{\mathcal{E}}}^{(k)}>$. Since the object part is moved by the robot in the interactive perception stage, the sub-part closest to the robot end effector $\bm{Q}_\text{closest}$ must be moved. Therefore, we traverse $\bm{{\mathcal{G}}}^{(k)}$ in a breadth-first order starting from $\bm{Q}_\text{closest}$ and add another criteria: sub-part $\bm{Q}$ is moved only if at least one predecessor sub-part of $\bm{Q}$ is moved. Algorithm~\ref{alg:movable_part_seg} illustrates the whole segmentation pipeline.

\begin{algorithm}

\renewcommand{\algorithmicrequire}{\textbf{Input:}}
\renewcommand{\algorithmicensure}{\textbf{Output:}}
\caption{{Movable Part Segmentation}}\label{alg:movable_part_seg}
\begin{algorithmic}[1]
\REQUIRE {List of meshes $\{\bm{{\mathcal{A}}}^{(k)}\}_{k=0}^{K}$}
\ENSURE {Movable part segmentation $\bm{{\mathcal{S}}}^{(K)}$ }
{
\FOR{$k=0,\dots,K$}
    \STATE Segment $\bm{{\mathcal{A}}}^{(k)}$ into small sub-parts $\bm{{\mathcal{Q}}}^{(k)}$ using SAMPart3D~\cite{yang2024sampart3d}
    \STATE Build $\bm{{\mathcal{G}}}^{(k)} = <\bm{{\mathcal{Q}}}^{(k)}, \bm{{\mathcal{E}}}^{(k)}>$ and find the sub-part closest to the robot end effector $\bm{Q}_\text{closest}$ 
    \STATE Sample point cloud $\bm{{\mathcal{P}}}^{(k)}$ with sub-parts indices from $\bm{{\mathcal{A}}}^{(k)}$
\ENDFOR
\STATE $\bm{{\mathcal{S}}}^{(0)}=0$ 
\FOR{$k=1,\dots,K$}
 \STATE \# Breadth-first traversal for graph $\bm{{\mathcal{G}}}^{(k)}$ starting from $\bm{Q}_\text{closest}$ 
    \FOR{Sub-part $\bm{Q}$}
    \STATE \# Determine whether each sub-part is moved
        \STATE Get point indices ${\bm{x}}$
        \STATE Sub-part point cloud $\bm{P}=\bm{{\mathcal{P}}}^{(k)}[\bm{x}]$
        \STATE Find $\bm{P}$'s closest points $\bm{PB}$ in $\bm{{\mathcal{P}}}^{(k-1)}$ and compute the distances $\{d_i^{(k)\rightarrow(k-1)}\}$
        \STATE Find $\bm{PB}$'s closest points $\bm{PF}$ in $\bm{{\mathcal{P}}}^{(k)}$ and compute the distances $\{d_i^{(k-1)\rightarrow(k)}\}$
        \STATE Compute whether $\bm{Q}$ is moved using $\{d_i^{(k)\rightarrow(k-1)}\}$ and $\{d_i^{(k-1)\rightarrow(k)}\}$
        \STATE Compute if at least one predecessor sub-part of $\bm{Q}$ is moved
        \IF{$\bm{Q}$ is moved}
        \STATE $\bm{{\mathcal{S}}}^{(k)}[\bm{x}]=k$ 
        \ELSE
        \STATE $\bm{{\mathcal{S}}}^{(k)}[\bm{x}]=\bm{{\mathcal{S}}}^{(k-1)}[\bm{x}_{\text{closest}}]$ 
        \ENDIF
    \ENDFOR
\ENDFOR
\RETURN $\bm{{\mathcal{S}}}^{(K)}$
}
\end{algorithmic}
\end{algorithm}

\subsubsection{Kinematic Structure Construction}
The final procedure is to construct the object's kinematic structure, including the part tree structure, joint categories (prismatic or revolute), joint configurations (axis direction and origin). Here we adopt ~\cite{liu2023building} to estimate the kinematic structure, where we first estimate the per-part rigid body transformations and further project them into a valid kinematic model. 

Finally, we transform the constructed model into the Unified Robot Description Format (URDF), which can be easily loaded into widely used multi-body physics simulators, such as SAPIEN~\cite{Xiang_2020_SAPIEN}. Since the geometries of the real-world object are usually complex, the reconstructed meshes can be non-convex. We further perform convex decomposition using VHACD~\cite{mamou2009simple} before importing the meshes to the physics simulator, which is essential for realistic physics simulation of robot interaction.
}

\subsection{Sampling-based Model Predictive Control}
\label{Sampling Based Model Predictive Control}
Having an explicit physics model and a target joint state $s_{target}$ of the articulated object, the agent needs to search for a trajectory that can change the current joint state $s_{initial}=s_{1}$ to $s_{target}$. The expected relative joint movement is $\Delta s_{target} = s_{target} - s_{initial}$. %
Because of the complex contact between the robot end effector and the constructed object model, the informative gradient of the objective function can hardly be acquired. Therefore, we employ sampling-based model predictive control, which is a zeroth-order method, to search for an optimal trajectory. There are various kinds of sampling-based model predictive control algorithms according to the zeroth-order optimization method used, such as  Covariance Matrix Adaptation Evolution Strategy (CMA-ES)~\cite{hansen2003reducing}, Cross-Entropy Method (CEM)~\cite{botev2013cross}, and Model Predictive Path Integral Control (MPPI)~\cite{williams2016aggressive}. Among these methods, we select the iCEM method \cite{pinneri2020sample} to search for a feasible long-horizon trajectory to complete the task due to its simplicity and effectiveness. We briefly describe how we apply the iCEM method in the following paragraph.

Trajectory length $T\in\mathbb{N}^+$ denotes the maximum time steps in a trajectory. 
At each time step $t (t<T)$, the action of the robot $\bm{a}_t\in \mathbb{R}^d$ is the incremental value of the joint position, where $d$ is the number of degrees of freedom (DOF) of the robot. The population denotes the number of samples sampled in each CEM iteration. Planning horizon $h$ determines the number of time steps the robot plans in the future at each time step. The top $E$ samples according to rewards compose an elite set, which is used to fit means and variances of a new Gaussian distribution. Please refer to~\cite{pinneri2020sample} for details of the algorithm.

At each time step $t$, the agent generates an action for the robot $\bm{a}_t\in \mathbb{R}^d$, where $d$ is the dimension of the action space. {For suction grippers, $d=7$, which is the DOF of the robot arm.} For 2-finger gripper tasks, $d=8$, which consists of the DOF of the robot arm and the 1 DOF of the gripper. However, for dexterous hands, $d=7+d_{\text{hand}}$, which includes the 7 DOF of the robot arm and the $d_\text{hand}$ DOF for the hand. The computational cost of iCEM is multiplied due to the high dimensionality of the action space. Consequently, directly searching in the original joint space of the multi-finger dexterous hand is not feasible. Moreover, the high-dimensional space of the dexterous hand may lead to unnatural postures. Therefore, it becomes essential to reduce the action space within the iCEM algorithm when using the dexterous hands.

\label{sec:method_eigengrasp}
In our work, we propose to use eigengrasp~\cite{ciocarlie2007dexterous} to reduce the high-dimensional action space associated with dexterous hands. This approach involves clustering a substantial number of grasping object actions to extract low-dimensional principal components. These components are then linearly combined to approximate the hand’s grasping posture. We first generate a dataset containing a diverse set of distinct grasping postures of the dexterous multi-finger hand in simulation using DexGraspNet~\cite{wang2023dexgraspnetlargescaleroboticdexterous}. 
We perform Principal Components Analysis (PCA)  on the dataset to obtain the first $m$ eigenvectors $\boldsymbol{e}_1$, ..., $\boldsymbol{e}_m$ with largest eigenvalue. 
The iCEM algorithm is then performed on the action space $\bm{a}_t\in \mathbb{R}^{7+m}$. The joint angles of the hand $\bm{q}_h$ are computed as a linear combination of the $m$ eigenvectors:

\begin{equation}
    \bm{q}_h= \sum_{i=1}^{m}{(a_i \cdot \bm{e}_i)}
\end{equation}

To speed up the search process, we use dense rewards to guide the trajectory optimization:
{
\subsubsection{Suction gripper}
For the suction gripper, the reward function consists of the following terms:

\noindent (1) \emph{success reward}
\begin{equation*}
            r_{success}=\begin{cases}\omega_s, \quad & \text{if} \left|s_{target}-s_t\right|<\epsilon\\
                                     0, \quad & \text{else}\end{cases}
        \end{equation*}
\noindent where $s_t$ denotes the joint state at current time step $t$, and $\epsilon$ is a predefined threshold.

\noindent (2) \emph{approaching reward}
\begin{equation*}
            r_{target} = -\omega_t \cdot \frac{s_{target}-s_{t}}{s_{target}-s_{initial}} 
        \end{equation*}
\noindent         This reward encourages $s_{t}$ to converge toward $s_{target}$.

\noindent (3) \emph{contact reward}
\begin{equation*}
            r_{contact} = 
            \begin{cases}
                    \omega_{contact}, & \text{if}\quad \dfrac{\left| s_{t}-s_{target} \right|}{\left| s_{target}-s_{initial} \right|}<1\\
                    -\omega_{collision}, & \text{if unexpected collision happens} \\
                    0, & \text{else}
            \end{cases}
        \end{equation*}
\noindent  This reward encourages the robot to establish initial contact with the object in the correct direction and maintain contact while moving the part. Additionally, it prevents collisions of parts other than the fingertip or the target part of the object.

\noindent (4) \emph{distance reward}
\begin{equation*}
            r_{dist} = \omega_d \cdot || \bm{p}_{virtual} - \bm{p}_{grasp} ||_2^2
        \end{equation*}
\noindent 
It is important to note that, unlike the distance reward for two-finger grippers or dexterous hands, which encourage the entire end effector to approach the target link of the object, the suction gripper focuses on bringing the tip of the suction closer to the target link. To achieve this, we mount a virtual link, $\bm{l}_{virtual}$, at the tip of the suction gripper for reward computation.
This reward encourages the suction gripper to move closer to the target part of the object, where $\bm{p}_{grasp}, \bm{p}_{virtual}\in \mathbb{R}^3$ represent the position of the geometry center of the target part and the position of the suction tip, respectively.

\noindent (5) \emph{access direction reward}
\begin{equation*}
            r_{dir} = \begin{cases}\omega_{dir},  \quad 
           \text{if}\quad \dfrac{ \mathbf{v}_{suction} 
\cdot \mathbf{n}_{target} }{|| \mathbf{v}_{suction} ||_2 \cdot || \mathbf{n}_{target}||_2}\geq\cos \psi\\
                                    
            0, \qquad\qquad\text{else}
                                     \end{cases}
        \end{equation*}
\noindent 
This reward function encourages the suction to make contact with the target link's surface within a certain range of directions. $\mathbf{v}_{suction}$ and $\mathbf{n}_{target}$ represent the orientation of the suction gripper and the normal of the target link's surface, respectively. $\psi$ is a predefined angle threshold.

}

\subsubsection{Two-finger gripper}

For the two-finger gripper, apart from 
\emph{success reward} $r_{success}$, \emph{approaching reward} $r_{target}$ and \emph{contact reward} $r_{contact}$ which remain consistent in the reward function of suction gripper, the other two terms are as follows:

\noindent (1) \emph{distance reward}
\begin{equation*}
            r_{dist} = \omega_d \cdot || \bm{p}_{part} - \bm{p}_{grasp} ||_2^2
        \end{equation*}
\noindent         This reward encourages the gripper to get closer to the target part of the object, where $\bm{p}_{part}, \bm{p}_{grasp} \in \mathbb{R}^3$ denotes the position of the geometry center of the target part and the grasp center of the gripper in Cartesian space, respectively.
        
\noindent (2) \emph{regularization reward}
 \begin{equation*}
            r_{reg} = -\sum_{i=1}^d{(\omega_a \cdot a_i + \omega_v \cdot v_i)}
        \end{equation*}
 \noindent        This reward is a regularization reward that discourages the robot to move too fast or move to an unreasonable configuration. $a_i$ and $v_i$ denote the acceleration and velocity of the $i$th joint respectively.

\subsubsection{Dexterous hand}
\label{sec:dex_hand_reward}

For the dexterous hand, apart from the \emph{success reward} $r_{success}$ and \emph{approaching reward} $r_{target}$, which remain consistent in the suction gripper’s reward function, the other three terms are as follows:

\noindent (1) \emph{contact reward}
\begin{equation*}
            r_{contact} = \begin{cases}\omega_{contact},  \quad 
            \text{if} \enspace \textbf{ISCONTACT}(palm,obj)\enspace\textbf{AND}\\
            \qquad\qquad\sum\limits_{finger}^{}\textbf{ISCONTACT}(finger,obj)\geq 2\\
            0, \qquad\qquad\text{else}
                                     \end{cases}
        \end{equation*}
\noindent         This reward function encourages the dexterous hand to cage much of the target link while searching for the trajectory.  With this reward, the dexterous hand can quickly find a stable grasping position of the target link and keep in contact with the object while moving the part.
        
\noindent (2) \emph{distance reward}
\begin{equation*}
            r_{dist} = \omega_d \cdot || \bm{p}_{part} - \bm{p}_{grasp} ||_2^2
        \end{equation*}
\noindent         This reward encourages the dexterous hand to get closer to the target part of the object, where $\bm{p}_{part}, \bm{p}_{grasp} \in \mathbb{R}^3$ denotes the position of the geometry center of the target part and the grasp center of dexterous hand, respectively.
        
\noindent (3) \emph{regularization reward}
 \begin{equation*}
            r_{reg} = -\omega_p \cdot e_p-\sum_{i=1}^d{(\omega_v 
\cdot v_i)}
        \end{equation*}
 \noindent        This reward discourages the robot to move too fast by restricting the joints' velocity. The reward also discourages position error of the end link using Cartesian error. $v_i$ denotes the velocity of the $i$-th joint respectively and $e_p$ denotes the Cartesian error of the end effector.

Once the manipulation trajectory is generated, we execute the trajectory on the real robot.

\makeatletter
\newcommand{\rmnum}[1]{\romannumeral #1}
\newcommand{\Rmnum}[1]{\expandafter\@slowromancap\romannumeral #1@}
\section{EXPERIMENTS}
\label{sec:experiments}
In this section, we evaluate the precision and effectiveness of the proposed method for manipulating articulated objects using {suction grippers,} two-finger grippers and dexterous hands.
We first conduct a large number of real-world articulated object manipulation experiments and quantitatively compare the performance. Then we design ablation studies to verify the effectiveness of different modules of our method, and validate the operational advantage of the dexterous hand against the two-finger gripper by comparing the task execution efficiency in simulation.
{Finally, we analyze the explicit world model construction accuracy and its impact on manipulation performance.}

\subsection{Experimental Setup}
\begin{figure}[t]
\centering
\subfigure[]{
\includegraphics[width=4cm]{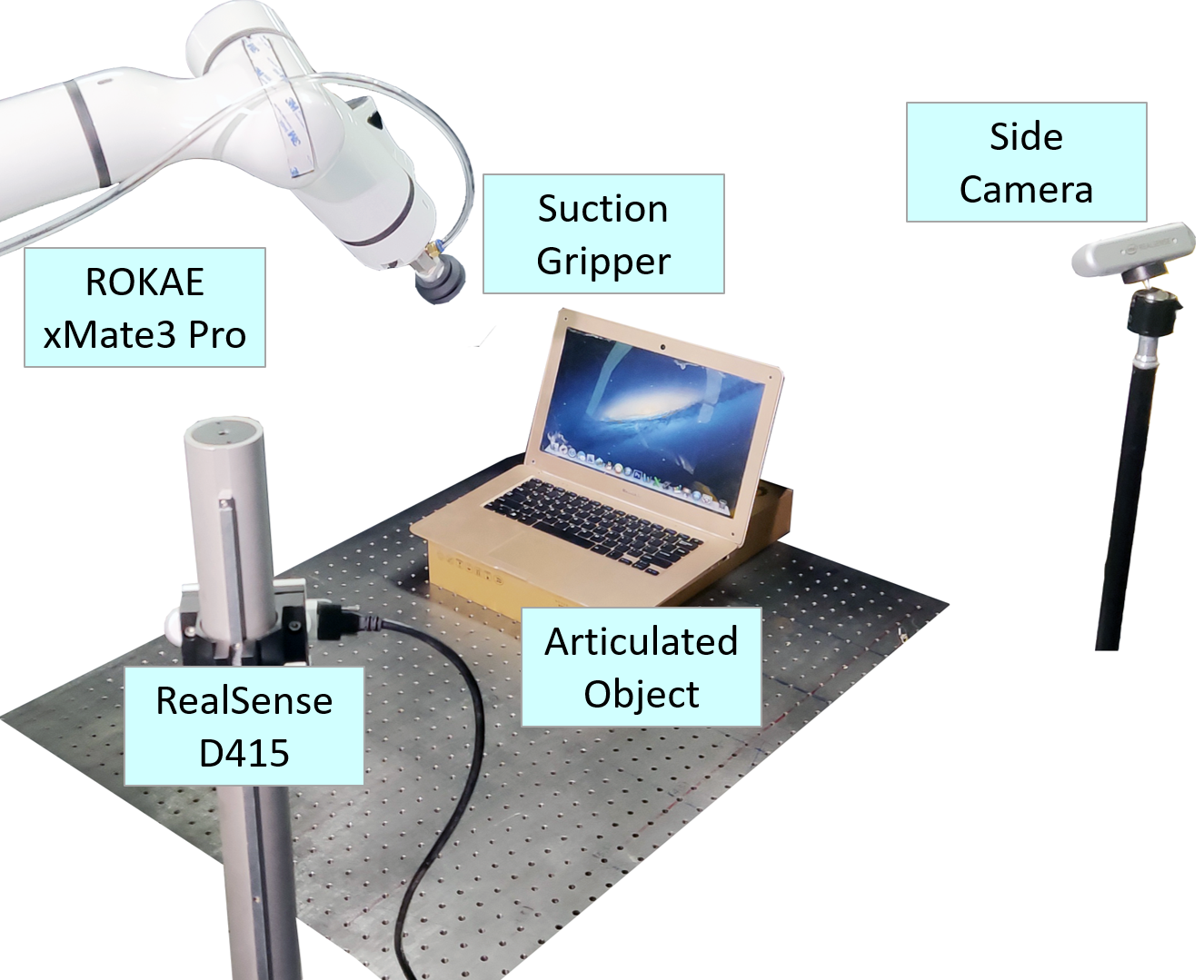}
}
\subfigure[]{
  \includegraphics[width=4cm]{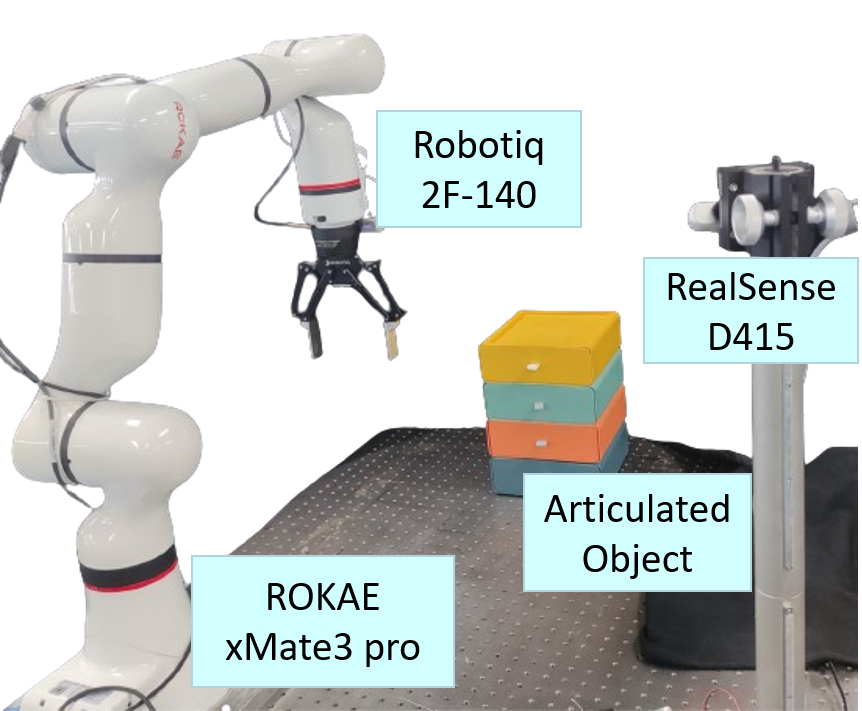 }
}
\hspace{-4mm}
 \subfigure[]
 {
  \includegraphics[width=4cm]{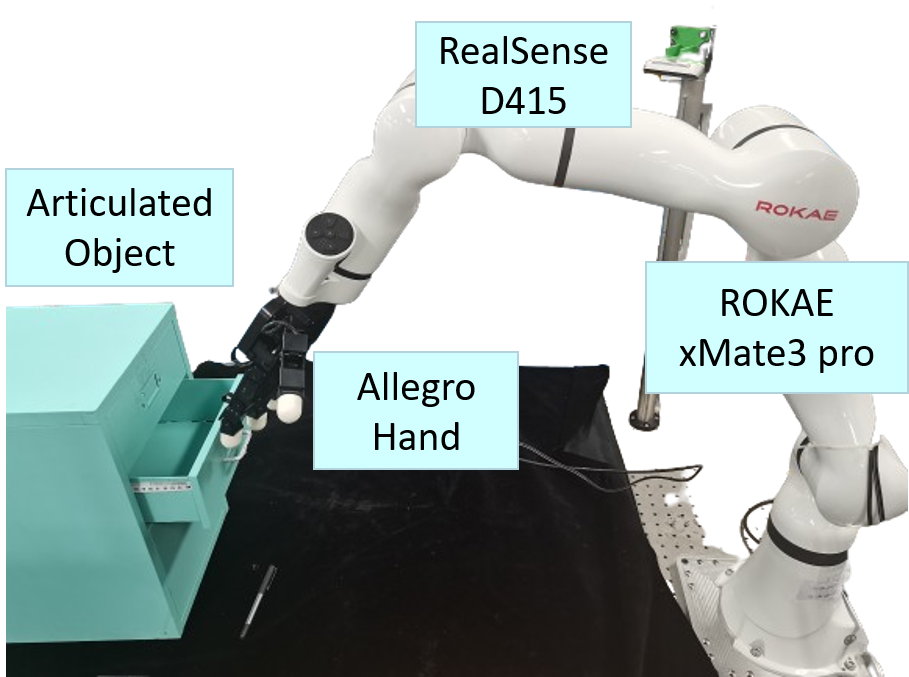 }
  \label{fig:setup1}
 }
  \hspace{-4mm}
\subfigure[]
 {
\includegraphics[width=4cm]{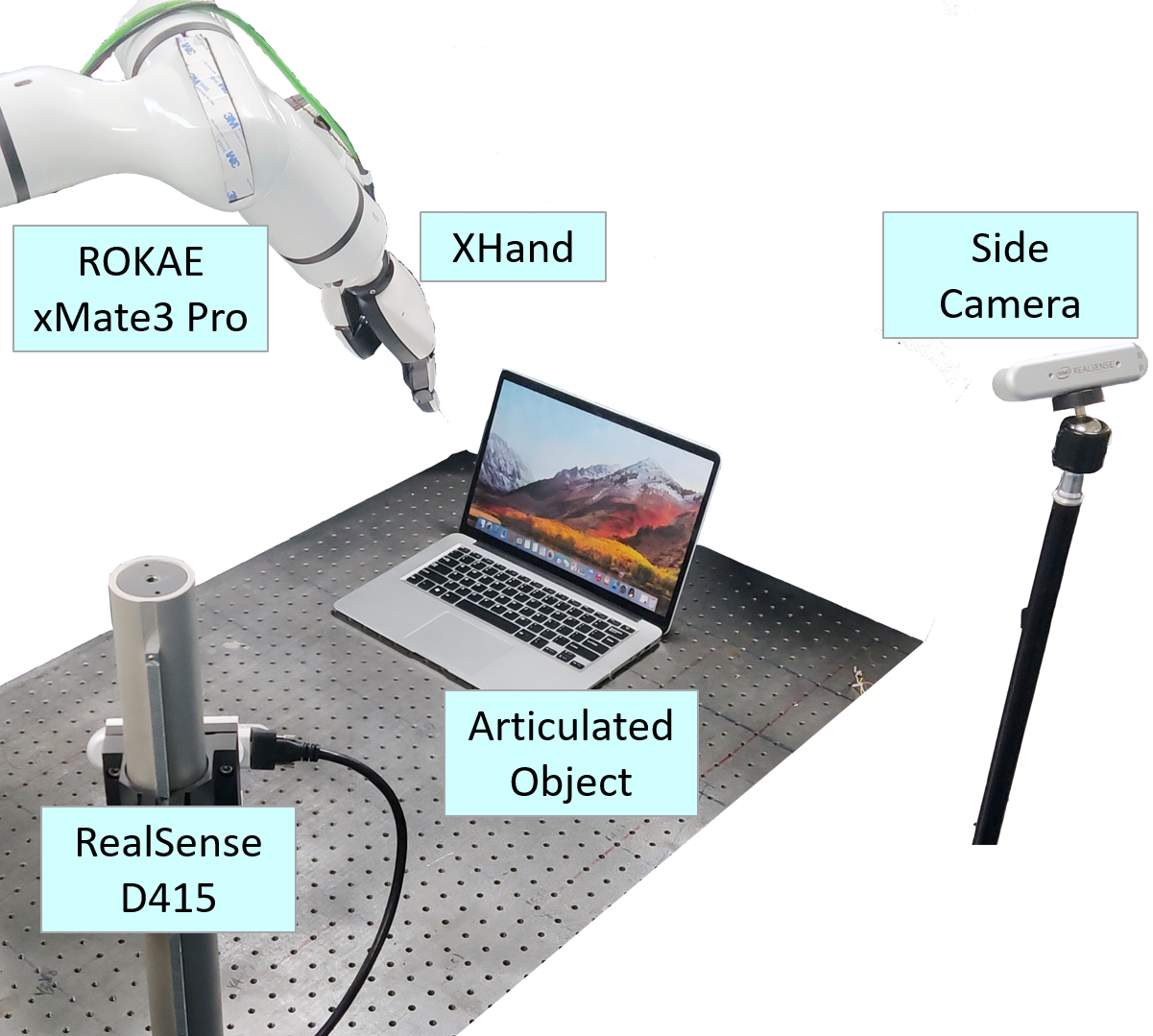}
  \label{fig:setup2}
 }
 \caption{{Real-world experimental setup of (a) suction gripper; (b) two-finger gripper and (c)(d)dexterous hands. Here we use another RealSense D415 as the Side camera.}}
 \label{fig:setup12}
\end{figure}

Fig.~\ref{fig:setup12} shows the real-world experimental setup. For the robot, a 7-DOF robot arm (ROKAE xMate3Pro) is used, which is fixed at the table. { An RGBD camera (Intel RealSense D415) is set to capture the visual input and  an additional RGB camera is positioned at a side viewpoint to capture side images that assist explicit model construction. Note that only the RGBD camera's intrinsic parameters and transformation relative to the robot base need to be calibrated, whereas no calibration is required for the RGB camera.} 
{4 kinds of end effectors are used: a suction gripper, a 1-DoF 2-finger gripper (Robotiq 2F-140), a 16-DoF 4-finger dexterous hand (Allegro Hand), a 12-DoF 5-finger dexterous hand (XHand).}

\begin{figure}
    \centering
    \includegraphics[width=6cm]{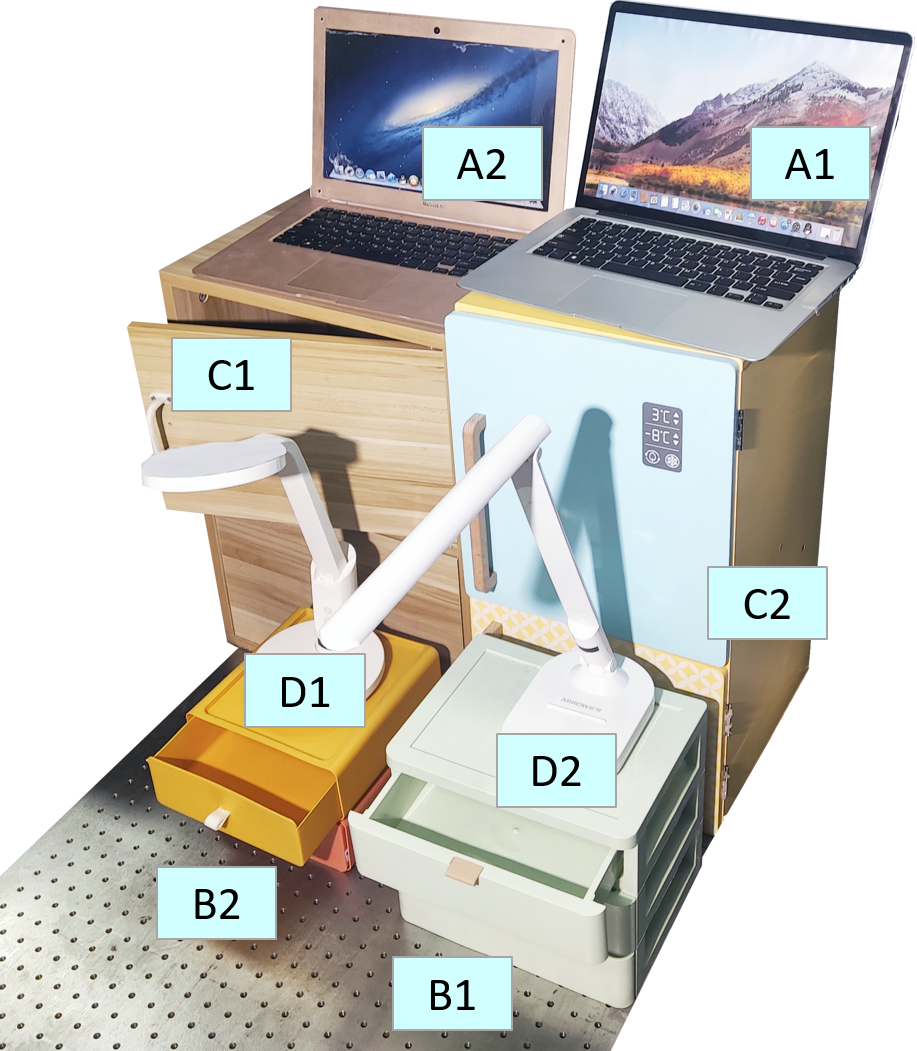}
    \caption{Articulated objects for manipulation.}
    \label{fig:objects}
\end{figure}

{We choose 4 categories of common articulated objects for experiments, which are drawers, cabinets, lamps, and laptops as shown in Fig.~\ref{fig:objects}. }

The articulated object is randomly located on the table with its base link fixed, and $s_{0}$ is randomly set. {We randomly select the goal state $\Delta s_{target}$ which does not exceed the joint limit and covers both directions of possible movement. We use a double-arm protractor ruler to measure the object's state after manipulation and compute the errors.
}

We use SAPIEN~\cite{Xiang_2020_SAPIEN} as the physics simulator to collect training data for the Explicit Physics Model Construction module and create simulation environments for the Sampling-based Model Predictive Control module.

\subsection{Implementation Details}

{
For affordance learning from self-supervised interaction, we introduce modifications to Where2Act~\cite{mo2021where2act} to accommodate our experimental setting: (1) we randomly set the target joint state; (2) the azimuth of the camera is randomly sampled from $[-60^\circ, +60^\circ)$ and the altitude is randomly sampled from $[15^\circ, 45^\circ)$; (3) we move the center of the point cloud to the origin. We downsample the object point cloud to 2000 points before feeding it into the network. For real-world deployment, we centralize the captured point cloud using an estimated bounding box.

For the 3D post-contact vector generation of VRB, we position the virtual camera at $(0.2, 0,  0.7)$ in the robot base frame, with a pitch angle of $\theta=52^\circ$. This configuration ensures a balanced between preserving visual features for the affordance network and ensuring sufficient offset for accurate computation of the 3D post-contact vector.
}

\begin{figure}[t]
      \centering
      \includegraphics[width=8.5cm]{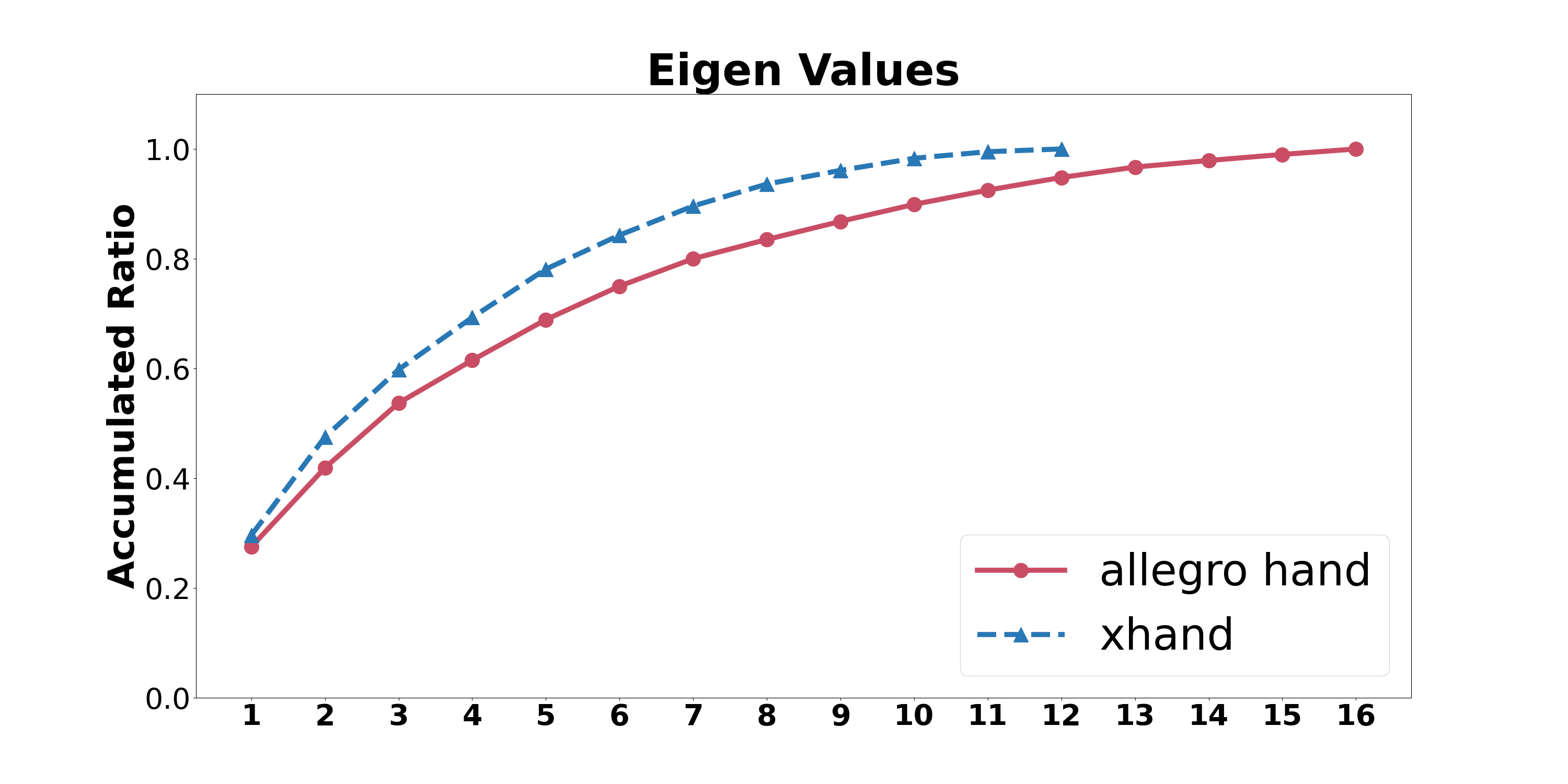}
      \caption{{Results of accumulated ratios of different eigengrasp dimensions for Allegro Hand and XHand.}}
      \label{fig:eigen_pca}
\end{figure}

To build the dataset for eigengrasp computation, we utilize DexGraspNet~\cite{wang2023dexgraspnetlargescaleroboticdexterous} to generate a collection of random grasping postures for the Allegro Hand and the XHand. The dataset includes 60800 grasp postures across 474 objects. We then compute the eigengrasp based on this data. Fig.~\ref{fig:eigen_pca} shows the accumulated ratios of different eigengrasp dimensions. {Interestingly, we find that the trends of the two hands, despite having different numbers of DoFs, are well aligned, suggesting that the eigengrasp effectively captures the fundamental features of the hand posture distribution. } Unless otherwise specified, we use eigengrasp dimension $m=2$ for dexterous manipulation experiments.

{
\subsection{Experiments on Suction Gripper}
\subsubsection{Implementation Details}
For parameters of the Sampling-based Model Predictive Control module, $T=50$, $E=20$, and $h=10$ are able to complete all the tasks. The range of incremental value of joint position is set to $[-0.05, 0.05]$. The parameters in the reward function are determined manually according to experience in the simulation environment. We set $\epsilon=0.005$ for prismatic joints, $\epsilon=0.02$ for revolute joints, $\omega_s=20$, $\omega_t=50$, $\omega_{contact}=10$, $\omega_{collision}=60$, $\omega_d=10$, $\omega_a=0.01$, $\omega_v=0.03$ and $\psi=15^\circ$. We use 30 processes for sampling in simulation on a computer that has an AMD Ryzen 9 9950X 16-Core CPU and an NVIDIA 3080Ti GPU. It takes about 1.5 minutes to find a feasible trajectory.

\subsubsection{Comparison With Baseline}
We choose UMPNet~\cite{xu2021umpnet} as the baseline method, as it can also achieve goal-conditioned manipulation. We choose two categories: Laptop and Cabinet and conduct 10 experiments for each category. After the trajectory is executed in the real world, we measure the real joint movement $\Delta s_{real}=s_{real}-s_{initial}$ and compare it with the target joint movement $\Delta s_{target}=s_{target}-s_{initial}$. We compute the error $\delta=\Delta s_{real}-\Delta s_{target}$ and the relative error $\delta_r={\delta}/{\Delta s_{target}}\times 100\%$.
Results of all the experiments can be found in Fig.~\ref{fig:real_experiement_results_suction}, and statistical results can be found in Table~\ref{tab:experimental_data_analysis_suction}. As shown, our method achieves significantly lower error compared to UMPNet. This improvement may be attributed to the fact that UMPNet uses a fixed step size, whereas our method employs continuous values in MPC. Additionally, the two images representing the current state and the goal state may not provide sufficient information for precise manipulation in UMPNet.

\begin{figure}[t]
      \centering
      \includegraphics[width=8.5cm]{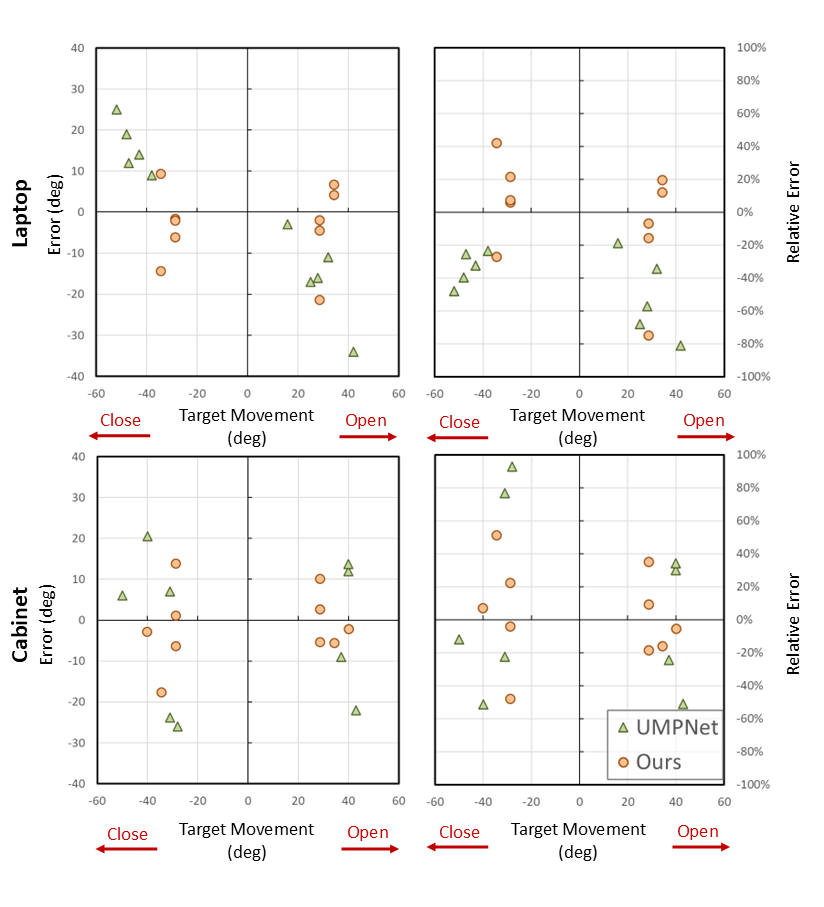}
      \caption{{Comparison of our method and UMPNet on real-world articulated object manipulation with suction gripper.}}
      \label{fig:real_experiement_results_suction}
\end{figure}

\begin{table}[thbp]
    \centering
    \caption{{Accuracy comparison of real articulated objects manipulation with suction gripper}}
    \resizebox{\columnwidth}{!}{%
    \begin{tabular}{c|cc|cc}

    \toprule
    {Category} &
     \multicolumn{2}{c|}{Laptop}&
     \multicolumn{2}{c}{Cabinet}\\
     \hline
     Method & UMPNet~\cite{xu2021umpnet} & Ours & UMPNet~\cite{xu2021umpnet} & Ours \\
     \hline
     $\textless$ 10\%  $\uparrow$ & 0/10 & \textbf{3/10}  & 0/10 & \textbf{4/10}   \\
     $\textless$ 30\%  $\uparrow$ & 3/10 & \textbf{9/10}  & 5/10 & \textbf{7/10}   \\
     $\textless$ 50\%  $\uparrow$ & 7/10 & \textbf{9/10}  & 6/10 & \textbf{8/10}   \\
      \hline
    {Avg $\left|\delta \right|$} $\downarrow$     & 16.0$^{\circ}$ & \textbf{7.2$^{\circ}$} & 15.3$^{\circ}$ & \textbf{6.8$^{\circ}$} \\
    {Avg $\left|\delta_r \right|$}$\downarrow$    & 42.9\% & \textbf{23.3\%} & 41.5\% & \textbf{21.7\%} \\ 
    \bottomrule
    
    \end{tabular}%
    }
    \label{tab:experimental_data_analysis_suction}
\end{table}

}

\subsection{Experiments on 2-Finger Gripper}
\subsubsection{Real World Articulated Object Manipulation}

For parameters of the Interactive Perception module, we choose $n_p=10$ and $n_a=10$.
For parameters of the Sampling-based Model Predictive Control module, we find that $T=50$, $E=300$, and $h=10$ are able to complete all the tasks. The range of incremental value of joint position is set to $[-0.05, 0.05]$. The parameters in the reward function are determined manually according to experience in the simulation environment. We set $\omega_s=20$, $\epsilon=0.005$ (m or rad), $\omega_t=50$, $\omega_{contact}=10$, $\omega_{collision}=60$, $\omega_d=10$, $\omega_a=0.01$ and $\omega_v=0.03$. We use 30 processes for sampling in simulation on a computer that has an AMD Ryzen 9 9950X 16-Core CPU and an NVIDIA 3080Ti GPU. It takes about 2.5 minutes to find a feasible trajectory.

\begin{figure}[t]
      \centering
      \includegraphics[width=8.5cm]{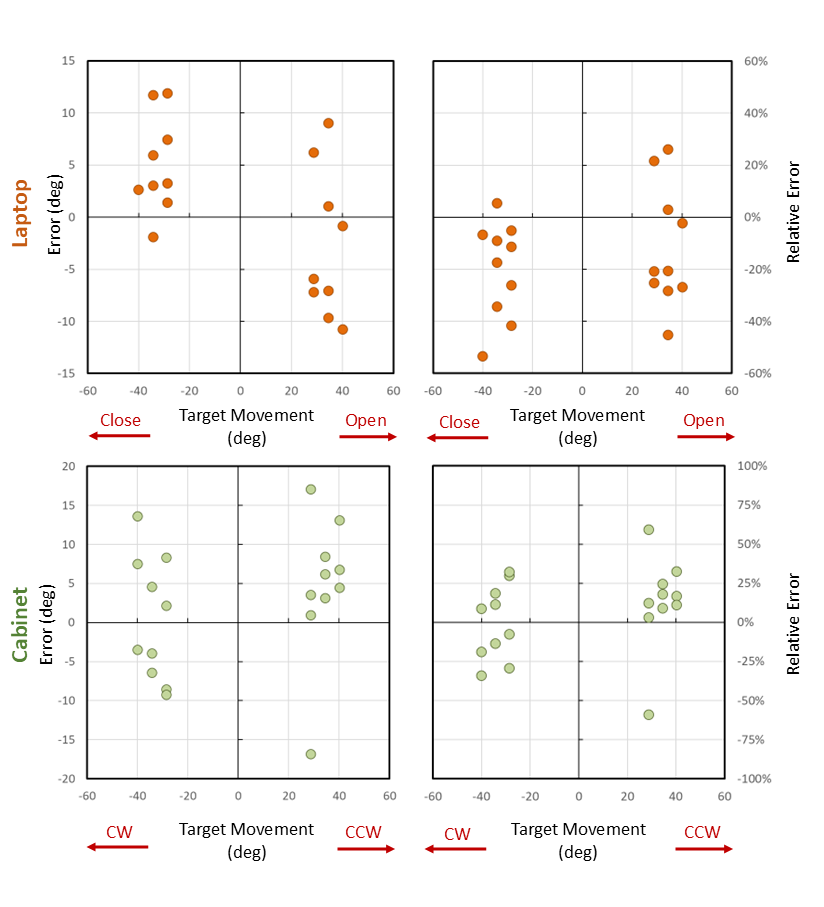}
      \caption{Experimental results of real-world articulated object manipulation with 2-finger gripper. Each row refers to the results of one category. The left column shows the error $\delta$ of the manipulation. The right column shows the relative error $\delta_r$ of the manipulation. The sign of the target movement denotes the direction of the movement (e.g. opening or closing, clockwise (CW) or counterclockwise (CCW).}
      \label{fig:real_experiement_results}
      
\end{figure}

\begin{figure}[h]
      \centering
    \subfigure[]
    {
    \centering
      \includegraphics[width=8cm]{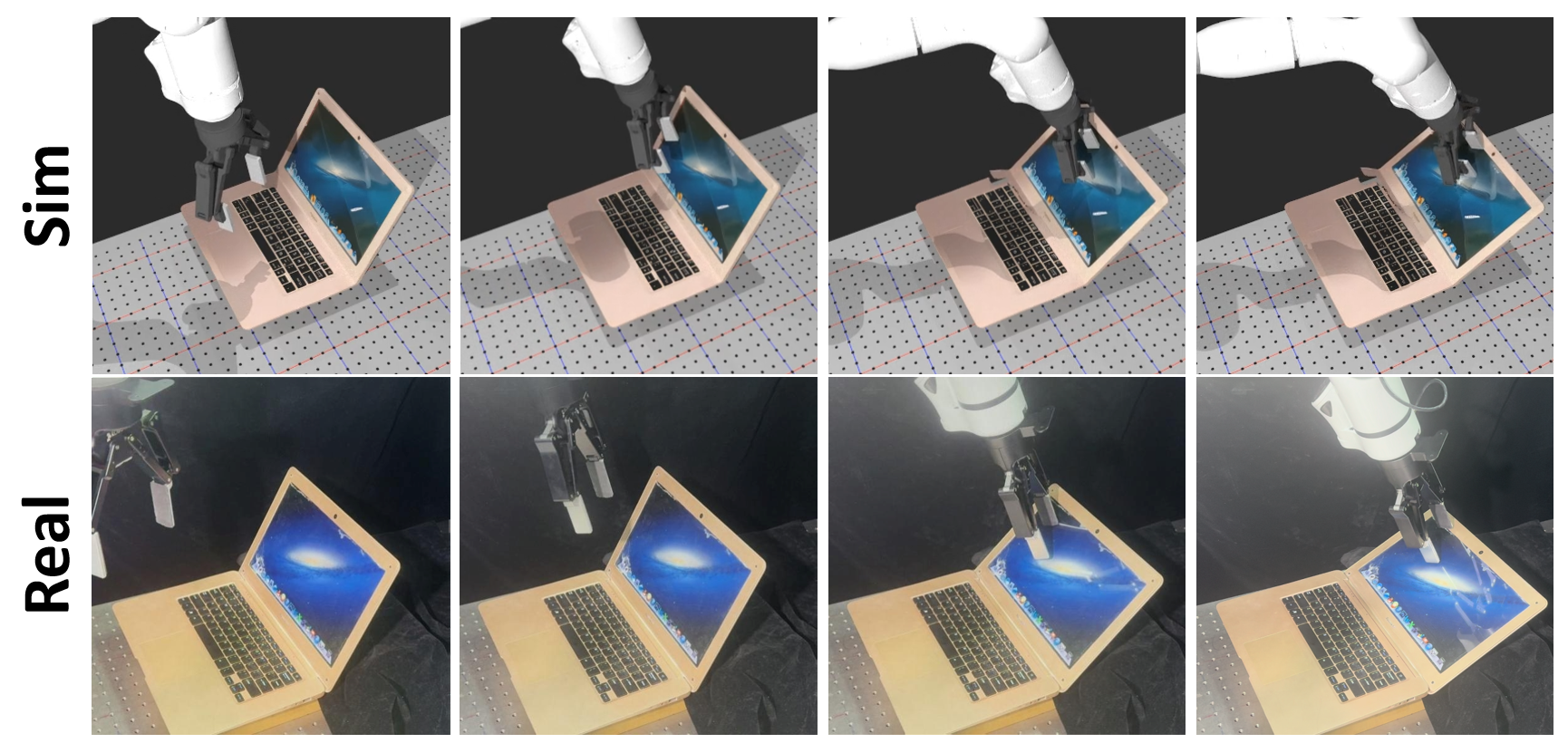}
        \label{fig:open_laptop}
     }
    \subfigure[]
    {
    \centering
      \includegraphics[width=8cm]{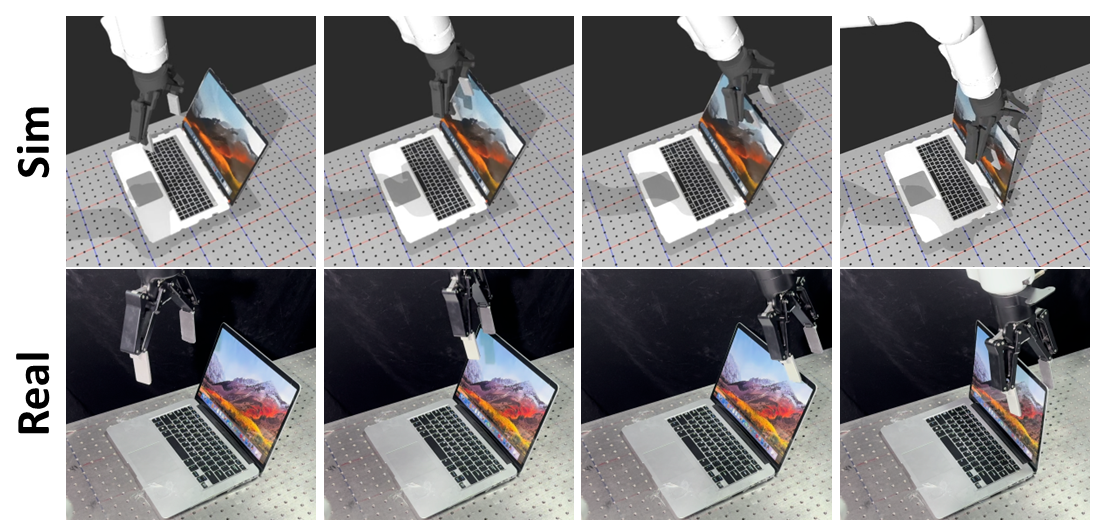}
        \label{fig:close_laptop}
      }
      \caption{Trajectories of laptop manipulation with 2-finger gripper: (a) open; (b) close. The constructed digital twin precisely captures the kinematic property of the articulated object, leading to the accurate alignment of Sim and Real.}
      \label{fig:trajectory_sequences_gripper}
\end{figure}

{We choose two categories: Laptop and Cabinet, and conduct 20 experiments for each category.
Results of all the experiments can be found in Fig.~\ref{fig:real_experiement_results}, and statistical results can be found in Table~\ref{tab:experimental data analysis}. Trajectories of both opening and closing the laptop are shown in Fig.~\ref{fig:trajectory_sequences_gripper}.
About $75\%$  of manipulations achieve a $\left|\delta_r \right| <30\%$, which shows the accuracy of our method.
}

\begin{table}[thbp]
    \centering
    \caption{{Accuracy of real articulated objects manipulation with 2-finger gripper}}
    \resizebox{\columnwidth}{!}{%
    \begin{tabular}{cc|cc}
    \toprule
    \multicolumn{2}{c|}{Category} &
      {Laptop} &
      {Cabinet}  \\ 
     \hline
    \multicolumn{2}{c|}{Number of manipulations} & 20   & 20    \\ \hline
    \multicolumn{1}{c|}{} & \textless 10\% &  6    &   4     \\
    \multicolumn{1}{c|}{} & \textless 30\% &  16    &   15   \\
    \multicolumn{1}{c|}{\multirow{-3}{*}{\begin{tabular}[c]{@{}c@{}}Number of manipulations \\ s.t. $\left|\delta_r \right|$ \end{tabular}}} &
      \textless 50\% & 20 &  18 \\ \hline
    \multicolumn{2}{c|}{Avg $\left|\delta \right|$}        & 7.2$^{\circ}$ & 7.4$^{\circ}$ \\
    \multicolumn{2}{c|}{Avg $\left|\delta_r \right|$}    & 21.5\% & 22.5\%   \\ \bottomrule
    \end{tabular}%
    }
    \label{tab:experimental data analysis}
\end{table}

\subsubsection{Ablation Study on Reward Function}
\begin{figure}[!thpb]
      \centering
      \includegraphics[width=8.5cm]{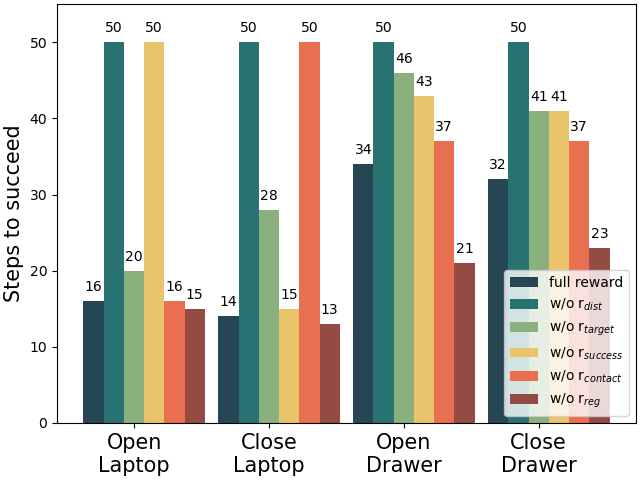}
      \caption{Results of ablation study on reward function. The 4 tasks are opening/closing drawer and opening/closing laptop. The task is considered to be failed if it is not done when time step reaches 50.}
      \label{fig:ablation_study}
      
\end{figure}

The reward function in the sampling-based model predictive control module is designed to guide the robot to complete the task. To examine the impact of each term of the reward function, we conduct the ablation study. There are 5 terms in the reward function, so 6 groups of experiments are conducted to reveal each term's influence against the full reward function. The first group runs iCEM with the full reward function as in Section~\ref{Sampling Based Model Predictive Control}. Each of the other 5 groups drops one term of the full reward function. In each group, 4 tasks are performed to make the results more general. The task that is considered to be failed if not completed within 50 time steps. Fig.~\ref{fig:ablation_study} summarizes the experimental results. 

The experiments using the full reward function are superior in both success rate and steps to succeed, except for the experiments without $r_{reg}$. However, the trajectories searched in w/o $r_{reg}$ are not suitable for real-world execution, because the robot tends to move to an unusual configuration which could be dangerous. Without $r_{dist}$, the robot cannot complete the task because the horizon is too short to achieve a positive reward. Omitting $r_{target}$, $r_{success}$, or $r_{contact}$ results in lower success rates, and even when successful, the robot requires more steps to complete the task.

\subsection{Experiments on Dexterous Hand}
\label{sec:experiments_dexterous}
\subsubsection{Real World Articulated Object Manipulation}

{We choose two categories: Drawer and Cabinet, and conduct 20 experiments for each category. Considering the FOV of the RGBD camera as well as the workspace of motion planning, we randomly set the location and initial joint status $s_0$ of articulated objects on the table in a certain range, such that the object is in the workspace of the manipulator. We randomly select $\Delta s_{target}$ which does not exceed the joint limit and covers both directions of possible movement.
}

We use the Allegro Hand to conduct all the quantitative experiments of dexterous hand. For parameters of Sampling-based Model Predictive Control module, we find that $T=50$, $E=100$, $h=10$ leads to fast searching as well as good performance. For the parameters in the reward function, we make adjustments based on the results of simulation experiments. We set $\omega_s=20$, $\omega_t=50$, $\omega_{contact}=10$, $\omega_d=10$, $\omega_c=0.001$ and $\omega_v=0.01$. 
We use eigen dimension $m=2$ to conduct real world manipulation. We use 30 processes for sampling in simulation on a computer that has an AMD Ryzen 9 9950X 16-Core CPU and an NVIDIA 3080Ti GPU. It takes about 3 minutes to find a feasible trajectory.

\begin{figure}[t]
      \centering
      \includegraphics[width=8.5cm]{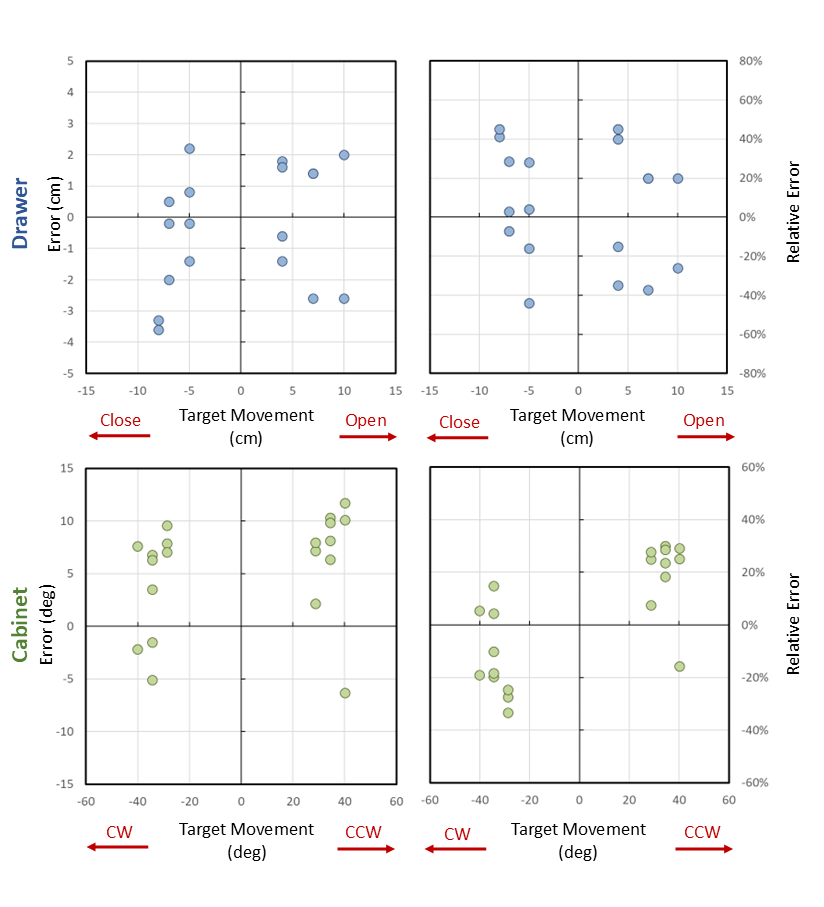}
      \caption{Experiment results of real-world articulated object manipulation with dexterous hands. Each row refers to the results of one category. The left column shows the error $\delta$ of the manipulation. The right column shows the relative error $\delta_r$ of the manipulation. The sign of the target movement denotes the direction of the movement (e.g. opening or closing, clockwise (CW) or counterclockwise (CCW).}
      \label{fig:real_experiement_results_hand}
      
\end{figure}

\begin{figure}[thpb]
      \centering

    \subfigure[]
    {
    \centering
      \includegraphics[width=9cm]{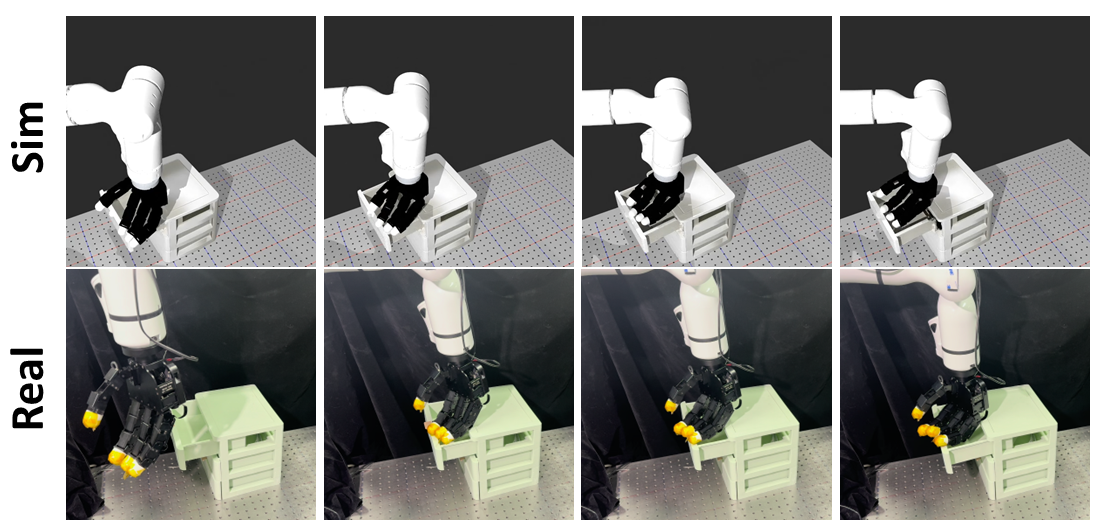}
        \label{fig:pull_drawer}
     }
     
    \subfigure[]
    {
    \centering
      \includegraphics[width=9cm]{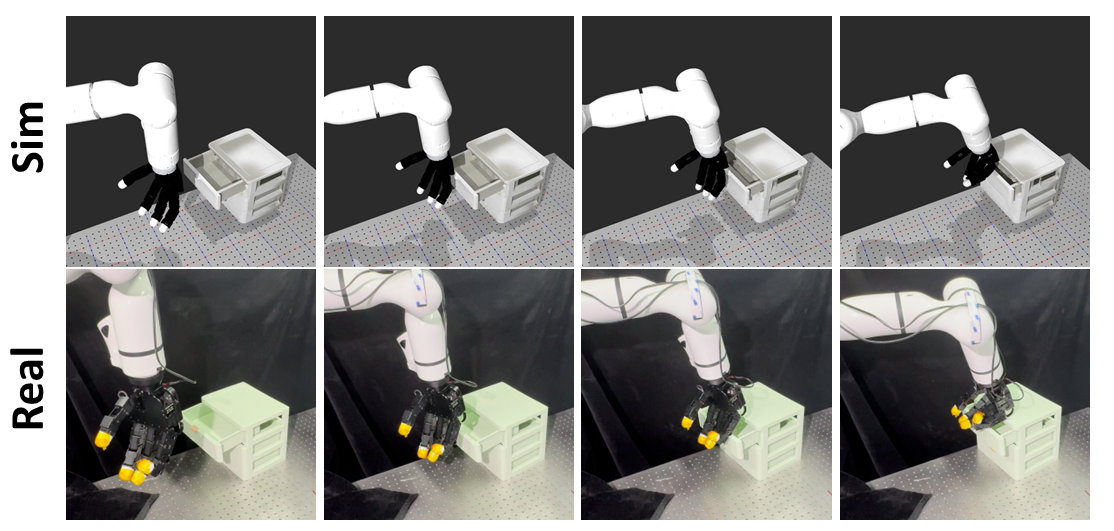}
        \label{fig:push_drawer}
      }
      \caption{Trajectories of drawer manipulation with dexterous hand: (a) open; (b) close. The constructed digital twin precisely captures the kinematic property of the articulated object, leading to the accurate alignment of Sim and Real.}
      \label{fig:trajectory_sequences_dexterous}
      
\end{figure}

Results of all the experiments can be found in Fig.~\ref{fig:real_experiement_results_hand}, and statistical results can be found in Table~\ref{tab:experimental data analysis hand}. Trajectories of opening and closing a drawer are shown in Fig.~\ref{fig:trajectory_sequences_dexterous}.

\begin{table}[thbp]
    \centering
    \caption{{Accuracy of real articulated objects manipulation with dexterous hand}}
    \resizebox{\columnwidth}{!}{%
    \begin{tabular}{cc|cc}
    \toprule
    \multicolumn{2}{c|}{Category} &
      {Drawer} &
      {Cabinet} \\ 
     \hline
    \multicolumn{2}{c|}{Number of manipulations} & 20   & 20  \\ \hline
    \multicolumn{1}{c|}{} & \textless 10\% & 3   &   3   \\
    \multicolumn{1}{c|}{} & \textless 30\% &  11   & 18  \\
    \multicolumn{1}{c|}{\multirow{-3}{*}{\begin{tabular}[c]{@{}c@{}}Number of manipulations \\ s.t. $\left|\delta_r \right|$ \end{tabular}}} &
      \textless 50\% & 20 & 20 \\ \hline
    \multicolumn{2}{c|}{Avg $\left|\delta \right|$}       &1.64cm  & 6.87$^{\circ}$ \\
    \multicolumn{2}{c|}{Avg $\left|\delta_r \right|$}    &26.4\%  &20.4\%
\\ \bottomrule
    \end{tabular}%
    }
    \label{tab:experimental data analysis hand}
\end{table}

\subsubsection{Ablation Studies} For the ablation study of dexterous hand manipulation, we investigate the influences of eigengrasp dimensions on Sampling-based Model Predictive Control the influences of different reward functions.

\paragraph{Eigengrasp Dimension}
In Section~\ref{sec:method_eigengrasp}, we propose to utilize eigengrasp to reduce the dimension of action space, enhancing the efficiency of our search process. 
To evaluate the effectiveness of MPC with different eigengrasp dimensions, we conduct experiments with dimensions $m=1,2,7,16$. We adopt 3 performance metrics: success rate, joint jerk, and computation time. 
We evaluate on 4 different tasks in simulation: opening/closing laptop and opening/closing drawer. Each task is repeated 10 times with randomized object positions and initial robot configurations.

\begin{enumerate}

\item ~\textbf{Success rate}
 Fig.\ref{fig:eigen_dim_success} presents the success rate of each task with different dimensions. The tasks of opening laptop and closing drawer achieve a 100\%  success across all the 4 dimension numbers. This high success rate likely stems from the simplicity of these tasks. 
 However, in the tasks of closing the laptop and opening the drawer, trajectories generated with $m=1$ shows a significantly lower success rate compared to the other dimensions. This may be attributed to the reduced dexterity of the Allegro Hand when operating in a 1-DOF configuration.
Interestingly, for all the 4 tasks, the dimension reduced to $m=2$ performs comparably to $m=7, 16$. This finding highlights the effectiveness of using eigengrasp space with $m=2$, achieving similar success rates while reducing computation time.
 
\begin{figure}[thpb]
      \centering
      \includegraphics[width=8.5cm]{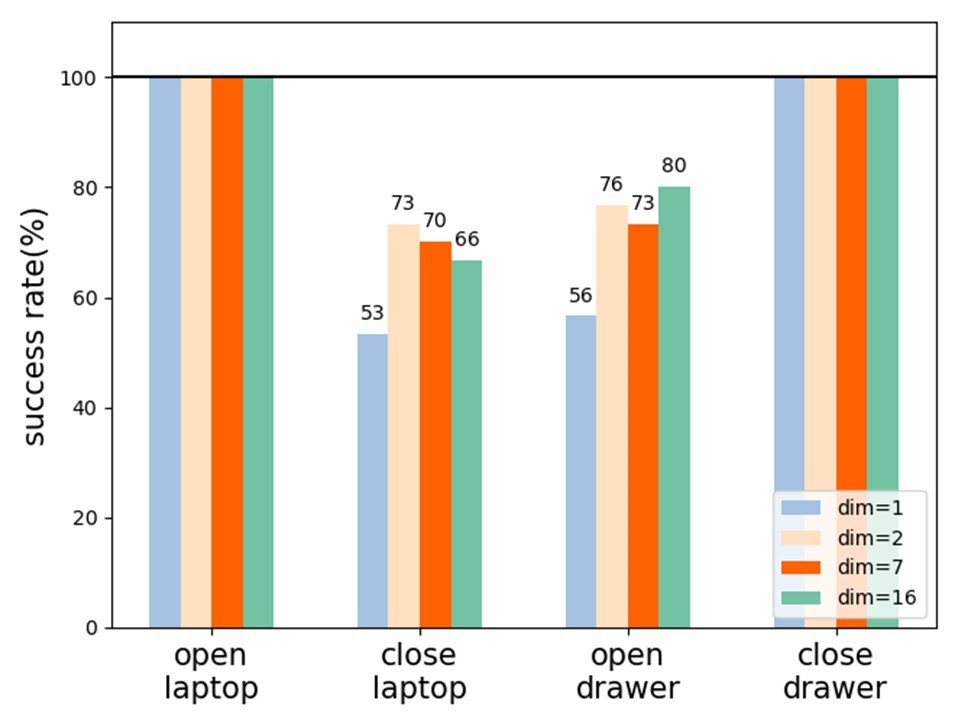}
      \caption{Success rate with different eigengrasp dimensions.}
      \label{fig:eigen_dim_success}
\end{figure}

\item ~\textbf{Joint jerk}
For real robot control, joint jerk reflects the smoothness of a robot's movements.  Our study investigates how the eigengrasp dimension affects the jerks of dexterous hand joints. To quantify jerk, we utilize the third derivative of hand joint positions with respect to time. Specifically, we calculate the average jerk for each finger at every time step, reflecting the smoothness of finger movement. Fig~\ref{fig:eigen_jerk} illustrates the results for the 4 tasks. 
Our experimental results consistently demonstrate that finger joint jerks increase with higher eigengrasp dimensions. Remarkably, for all the 4 tasks, dimension $m=1$ exhibits superior performance in terms of thumb jerk compared to $m=2,7,16$. However, for the remaining three fingers: both $m=1,2$ demonstrate similar advantages over higher dimensions.

\begin{figure}[thpb]
      \centering
      \includegraphics[width=8.5cm]{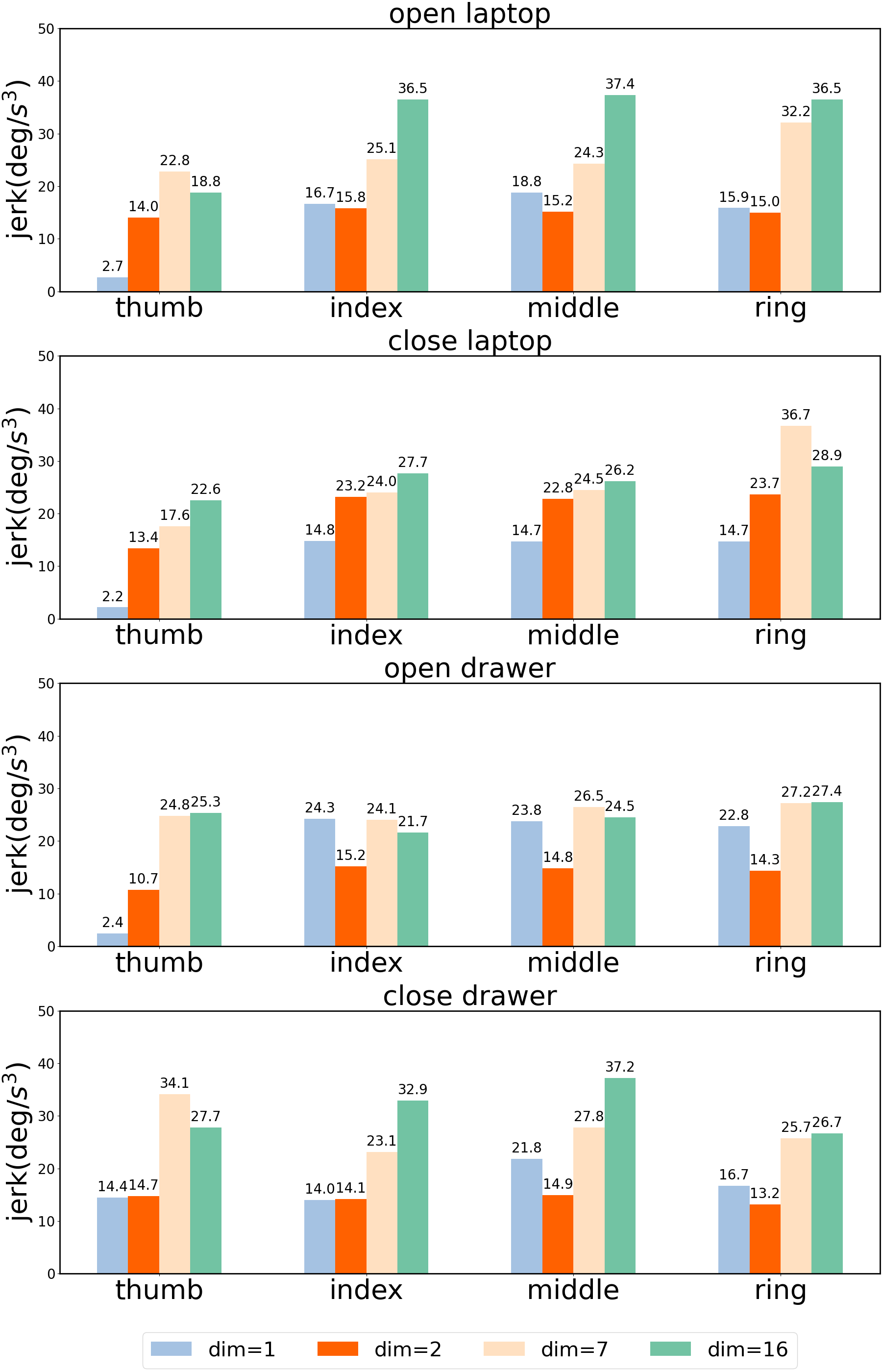}
      \caption{Joint jerk with different eigengrasp dimensions.}
      \label{fig:eigen_jerk}
      
\end{figure}

\item ~\textbf{Computation time}
For each task, we generate 30 different trajectories and compare the average time per step as shown in Fig~\ref{fig:eigen_dim_time}. 
It is shown that using eigengrasp dimension $m=2$ results in approximately 1 second less per step compared to 16 dimensions. Consequently, it takes nearly 1 minute less to find a feasible trajectory.

\begin{figure}[thpb]
      \centering
      \includegraphics[width=8.5cm]{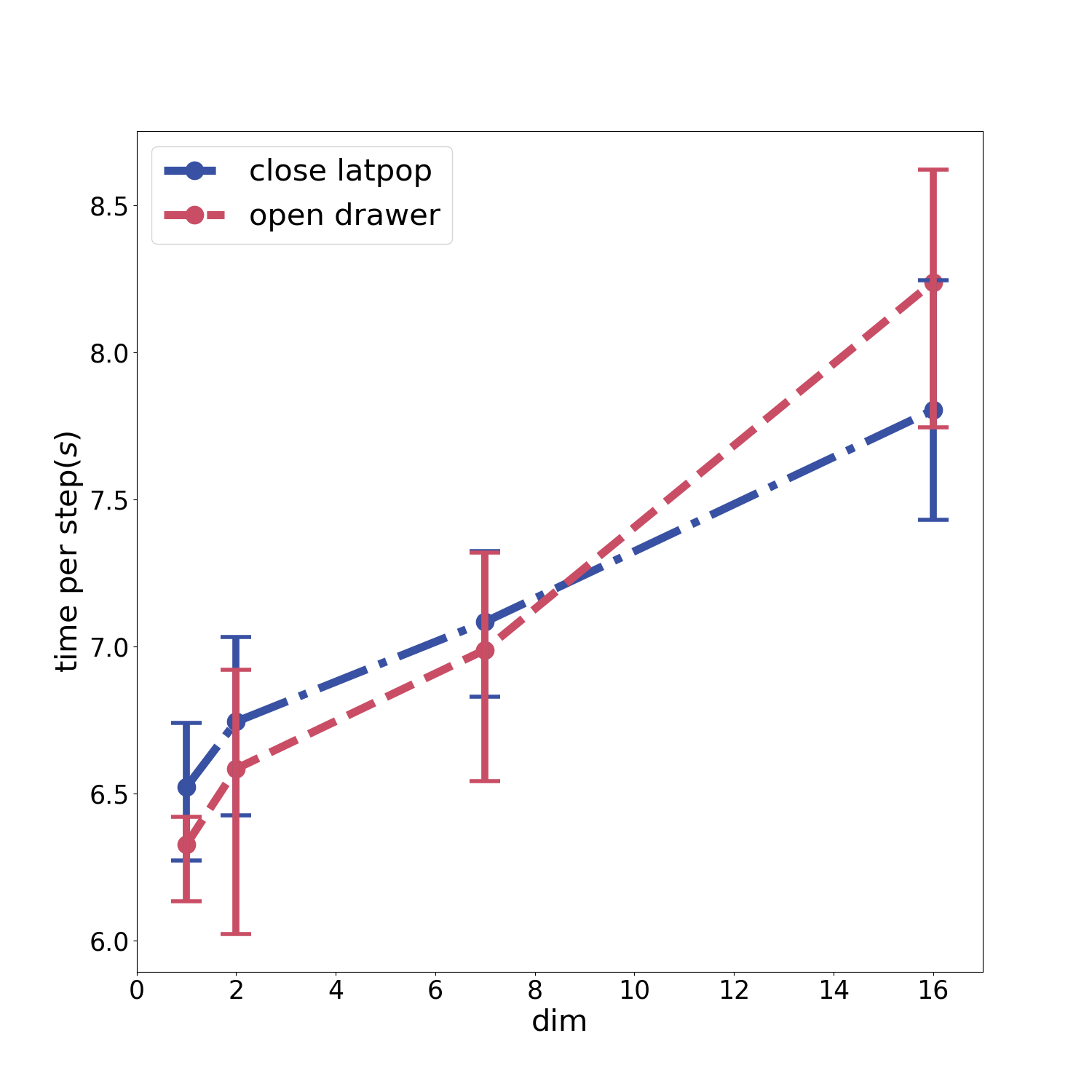}
      \caption{Computation time with different eigengrasp dimensions.}
      \label{fig:eigen_dim_time}
      
\end{figure}

\end{enumerate}

\begin{figure}[t]
      \centering
      \subfigure[]
      {
        \includegraphics[width=8cm]{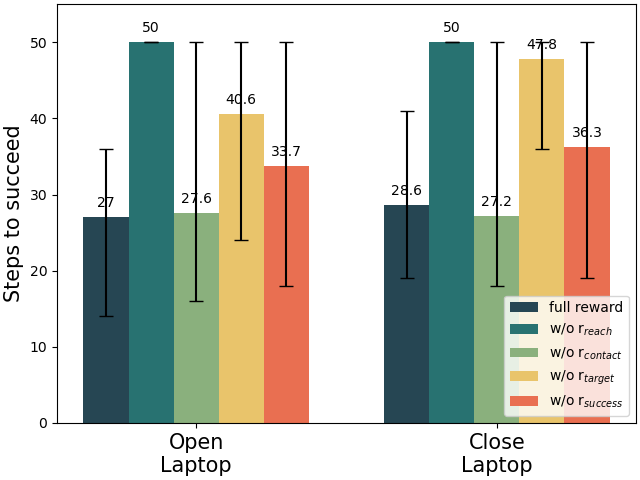 }
        \label{fig:dex_reward_steps}
      }
     
     \hspace{-4mm}
     \subfigure[]
     {
       \includegraphics[width=8cm]{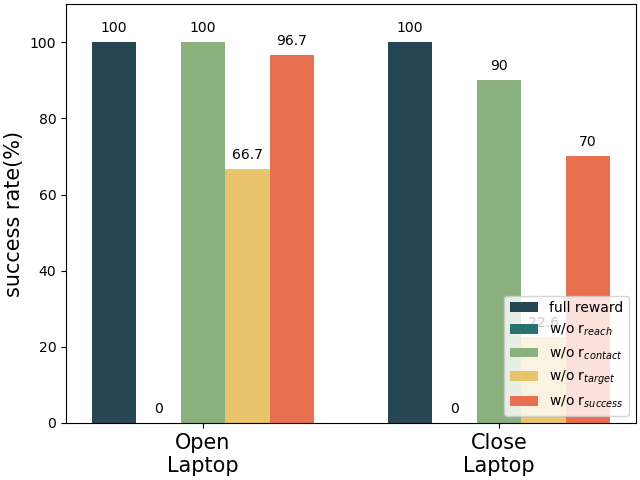}
       \label{fig:dex_reward_rate}
     }
     
     \caption{Results of numbers of steps to succeed (a) and results of success rate (b). The task is considered to be failed if it is not done when time step reaches 50. }
     
\end{figure}

\paragraph{Reward Function}
In Section~\ref{sec:dex_hand_reward}, we design reward function in the sampling-based model predictive control module for dexterous hand manipulation tasks. Five terms of reward functions are designed, which include $r_{success}$, $r_{target}$, $r_{contact}$, $r_{dist}$ and $r_{reg}$. We conduct 5 groups of experiments to reveal each term's influence against the full reward function. The first group runs iCEM with the full reward function as in~\ref{sec:dex_hand_reward}. The other 4 experiments drop one term of the full reward function. To make the ablation result more generalizable, we conduct 2 tasks for each group: opening laptop and closing laptop. In each task, we randomize the position of the object, the initial joint angle of the robot as well as the target qpos of the object to generate 30 different trajectory running iCEM. The tasks that are not done when time step reaches 50 are considered failed. Fig.~\ref{fig:dex_reward_steps} and Fig.~\ref{fig:dex_reward_rate} respectively summarize the experimental results.

The experiments using the full reward function consistently outperforms others in terms of both success rate and steps in completion, except for the experiments without $r_{contact}$. It's worth noticing that the reward function without $r_{contact}$ even  exhibits a surprising advantage in terms of the number of steps to success in task 2. This unexpected result may be attributed to the absence of constraints imposed by the human-like hand posture encouraged by $r_{contact}$. Without this component, the iCEM algorithm might explore unconventional hand postures to interact with the object. On the other hand, omitting $r_{dist}$ from the reward function makes the tasks impossible to accomplish for the robot. The short planning horizon prevents the robot from accumulating positive rewards. Similarly, excluding $r_{target}$ and $r_{success}$ leads to decreased success rates. In successful cases, the robot requires additional steps to accomplish the task.

\subsubsection{Advantage of Dexterous Manipulation}
In this section, we validate the advantages of the dexterous hand over the two-finger gripper through experiments on five tasks. For each task, we randomize the object's position and the robot's initial configuration 10 times. We then run the iCEM algorithm using both the Allegro hand and the Robotiq gripper. We use the number of steps to complete the task as the metrics. 

\begin{figure}[t]
      \centering
      \includegraphics[width=8.5cm]{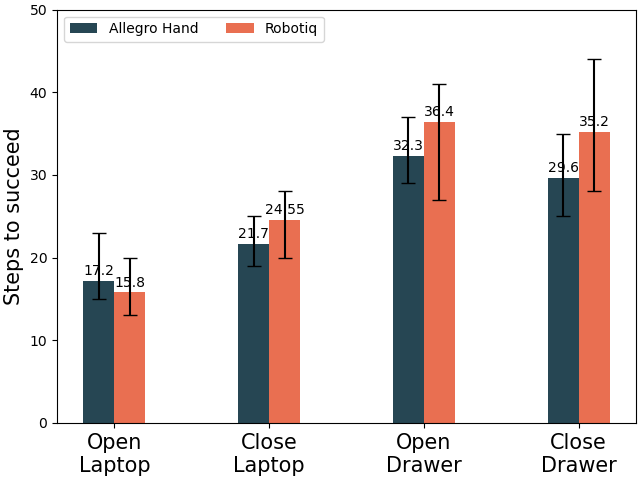}
      \caption{Comparison of manipulation with dexterous hand and two-finger gripper.}
      \label{fig:hand_gripper}
      
\end{figure}

\begin{figure}[t]
      \centering
      \includegraphics[width=8.8cm]{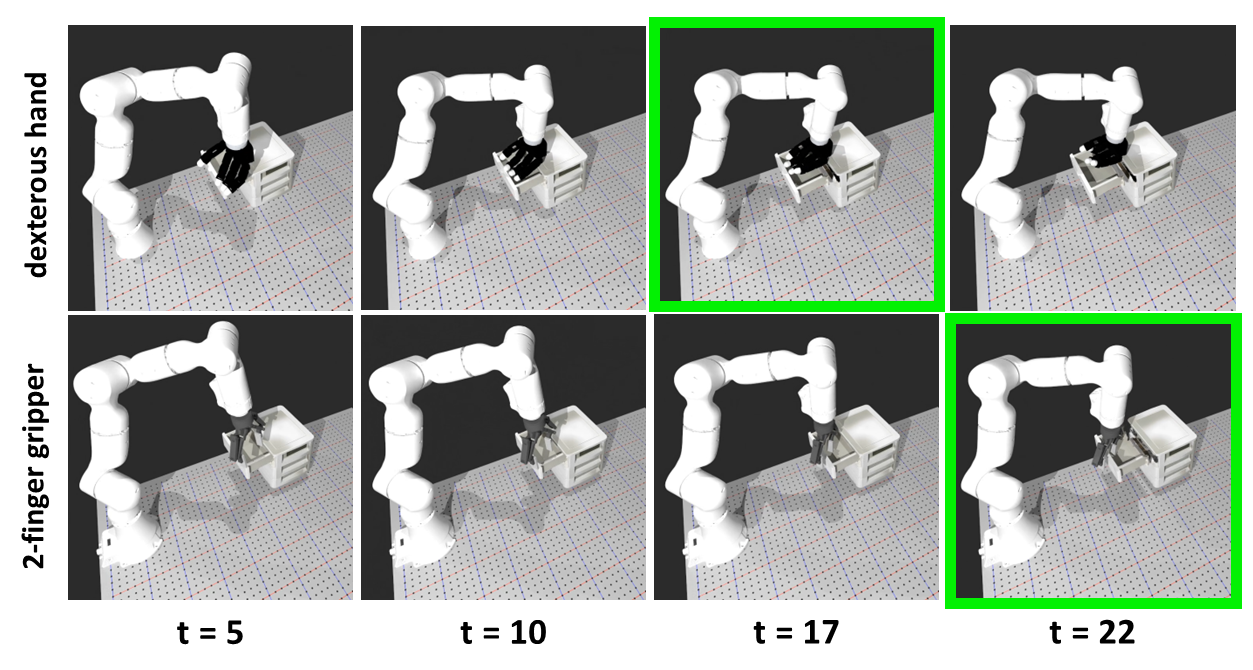}
      \caption{{Trajectory comparison of a dexterous hand and a 2-finger gripper performing the same drawer opening task. The dexterous hand completes the task in fewer steps (17 vs. 22).}}
      \label{fig:hand_gripper_step}
      
\end{figure}

Fig.~\ref{fig:hand_gripper} summarizes the comparison results between the dexterous hand and the two-finger gripper. Except for the laptop opening task, the dexterous hand consistently requires fewer steps on average. The anomaly in the laptop opening task can be attributed to its simplicity, as it does not require precise contact between the end effector and the object. Fig.~\ref{fig:hand_gripper_step} visualizes the trajectories for the laptop closing task, showing that our method is able to find a shorter trajectory for the dexterous hand by utilizing its additional degrees of freedom to close the laptop efficiently.

{
\subsection{Experiments on Explicit World Model Construction}
\subsubsection{Model Accuracy Evaluation}
{In this section, we quantitatively evaluate the model construction accuracy and compare it with ScrewNet~\cite{jain2021screwnet}.}
We use 4 metrics: RDE (Relative Depth Error), which is the relative error between the rendered depth map of the constructed model and the real depth map  captured by the RGBD camera; IoU (Intersection over Union), which measures the degree of alignment between the rendered mask and the groundtruth mask; Angle Err and Pos Err, which measure the joint axis error. For prismatic joints, only the angle error is applicable. 
{Since the training labels and the predictions of ScrewNet are defined in the object frame, which is determined by the CAD model's frame definition, we only evaluate the angle error of the joint axis for ScrewNet.}

We conduct experiments on all the four categories and collect 10 pairs of frames for each category.
Table~\ref{tab:model_accuracy} presents the evaluation results. Our method demonstrates the ability to accurately construct the world models across all four categories, achieving an average angle error of less than $9^{\circ}$. The relatively low IoU for lamps can be attributed to the small area the lamp occupies in the image, making it more sensitive to reconstruction errors.

\begin{table}[thbp]
    \centering
    \caption{{Quantitative evaluation of explicit world model construction accuracy}}
    \resizebox{\columnwidth}{!}{%
    \begin{tabular}{c|ccccc}
    \toprule
    \multirow{2}{*}{Category} & RDE & \multirow{2}{*}{IoU} & Pos Err & Ang Err & { Ang Err of} \\
     &  (\%) & &(m) & ($^{\circ}$) & {ScrewNet~\cite{jain2021screwnet} ($^{\circ}$)}\\
     \hline
     Laptop & 1.47 & 0.92 & 0.022& 1.34 & {23.15}  \\
    Cabinet & 3.77 & 0.94 & 0.029& 4.84 & {11.67} \\
    Drawer & 2.23 & 0.94  & -& 8.58 & {23.21} \\
    Lamp & 2.12 & 0.80 & 0.009 & 7.95 & {9.38}\\
    \bottomrule
    \end{tabular}%
    }
    \label{tab:model_accuracy}
\end{table}

\subsubsection{Impact of Model Construction Accuracy on Manipulation Performance}
In this section, we evaluate the impact of model construction accuracy on manipulation performance. We introduce different scales of joint position errors and joint orientation errors into the constructed model, search for the trajectories, and execute these trajectories on the real robot. Fig.~\ref{fig:modal_error_on_manipulate} shows the manipulation results. All the experiments are conducted on Laptop. As can be seen, the manipulation error is positively correlated with the model construction error, with position errors exerting a more significant impact.

\begin{figure}[t]
    \centering

    \includegraphics[width=\linewidth]{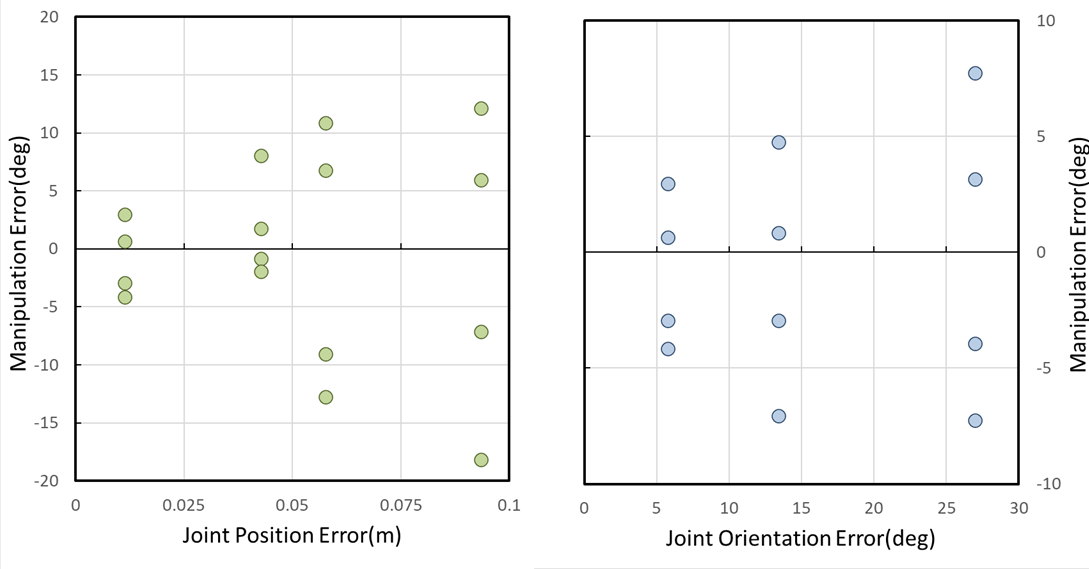 }
    \caption{{Real-world manipulation results of model construction with different scales of joint position errors and joint orientation errors.}}
    \label{fig:modal_error_on_manipulate}
    
\end{figure}

\subsubsection{Ablation Study on Movable Part Segmentation}

In this section, we validate the impact of movable part segmentation on kinematic structure construction. In \cite{liu2023building}, segmentation is jointly optimized with part motions using a distance loss. However, it may produce incorrect segmentation results when only two or three frames of point clouds are used. To address this, our proposed pipeline first segments the meshes and then estimates the kinematic structure. Table \ref{tab:model_accuracy_comp} compares the results across all four categories. Our method achieves more accurate segmentation and improved axis parameter estimation compared to \cite{liu2023building}, particularly for prismatic joints. Fig.~\ref{fig:segmentation_ablation} provides a visualization of some results. \cite{liu2023building} may generate an incorrect number of parts and wrong joint types.
 
\begin{table}[thbp]
    \centering
    \caption{{Comparison of model construction accuracy with and without our movable part segmentation.}}
    \resizebox{\columnwidth}{!}{%
    \begin{tabular}{c|c|ccc}
    \toprule
     Category & &Success &  Angle Err ($^{\circ}$) $\downarrow$& Pos Err (m) $\downarrow$\\ 
     \hline
     \multirow{2}{*}{Laptop} & w/ Part & \textbf{10/10}& 1.34 & \textbf{0.022} \\
      & w/o Part & 9/10 & \textbf{1.32} & 0.048\\
      \hline
     \multirow{2}{*}{Cabinet} & w/ Part & \textbf{10/10}& \textbf{4.84} & \textbf{0.029} \\
      & w/o Part & \textbf{10/10} & 6.10 & 0.034 \\
      \hline
     \multirow{2}{*}{Drawer} & w/ Part & \textbf{10/10}&  \textbf{8.58} & -\\
      & w/o Part & 5/10 & 16.04 & - \\
      \hline
     \multirow{2}{*}{Lamp} & w/ Part & \textbf{10/10}& 7.95 & \textbf{0.009} \\
      & w/o Part & 9/10 & \textbf{7.54} & 0.015\\

    \bottomrule
    \end{tabular}%
    }
    \label{tab:model_accuracy_comp}
\end{table}

\begin{figure}[t]
    \centering
        \subfigure[]{
            \includegraphics[width=4cm]{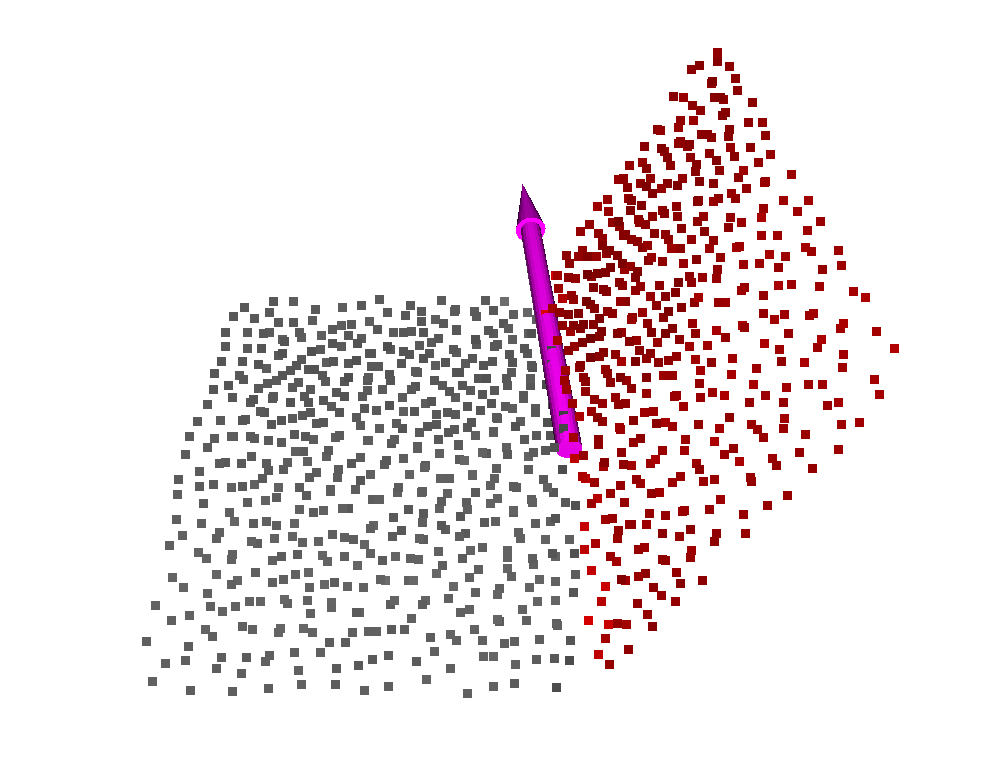}
        }
        \subfigure[]{
            \includegraphics[width=4cm]{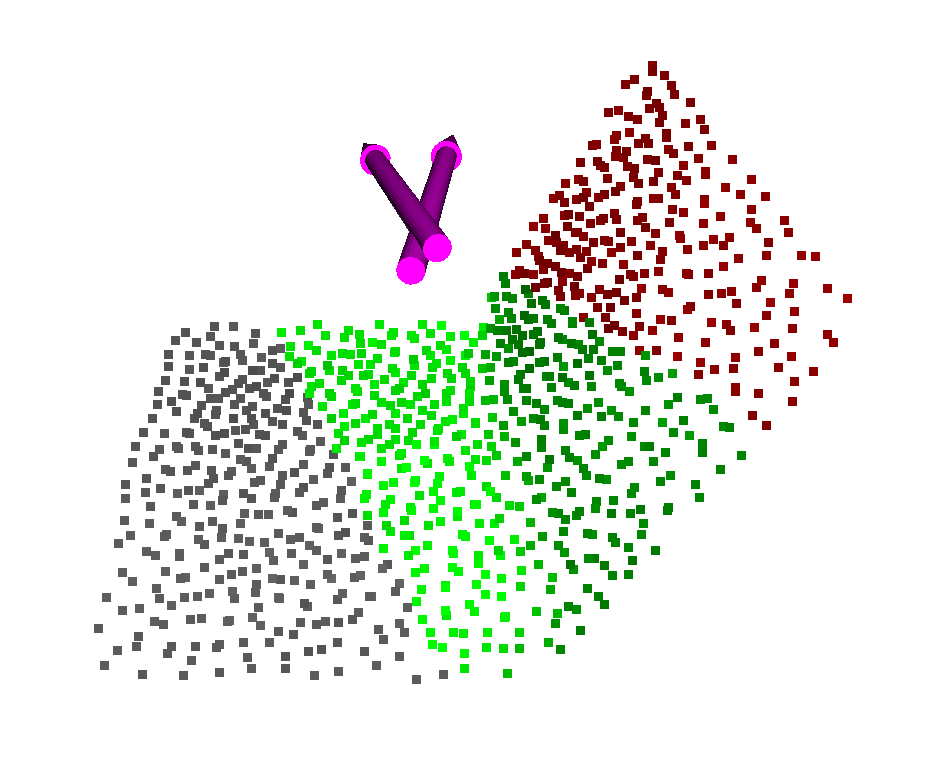}
        }
        
    \centering
        \subfigure[]{
            \includegraphics[width=4cm]{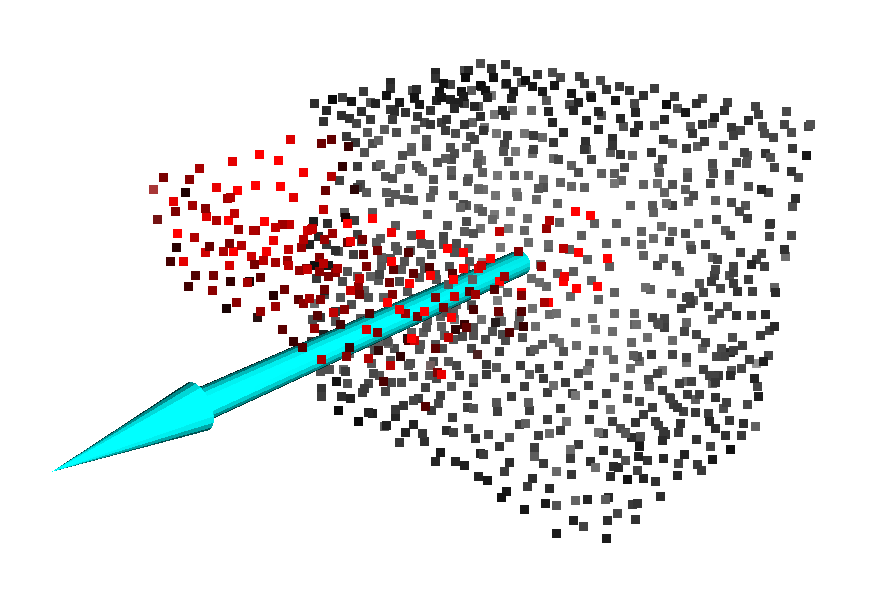}
        }
        \subfigure[]{
            \includegraphics[width=4cm]{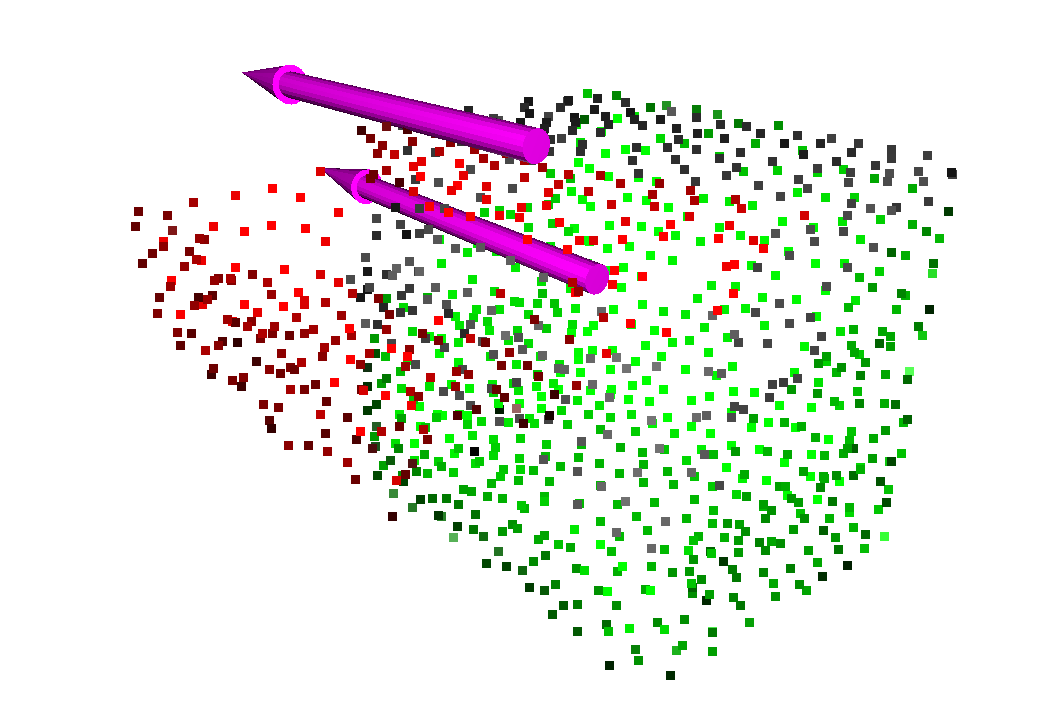}
        }

     \caption{{Visualization of model construction results: (a)(c)with our movable part segmentation, (b)(d)without our movable part segmentation. Revolute joints are shown in pink, and prismatic joints are shown in blue.}}
     \label{fig:segmentation_ablation}
\end{figure}
}

{
\subsection{Ablation Studies on Affordance Detection}
\subsubsection{Where2Act VS. VRB}
Fig.~\ref{fig:vrb_where2act} shows the affordance detection results of Where2Act and VRB across four categories. Since the goal of the one-step interaction is to change the state of the movable part, a pushing interaction is sufficient in most cases. For objects where pushing is not feasible—such as fully closed cabinets and drawers, the affordance detection module must detect the handles. For cabinets, VRB can detect handles in 17 out of 20 samples, while Where2Act can detect 7. For drawers, VRB can detect handles in 3 out of 20 samples, and Where2Act can only detect 1.  VRB achieves a much higher recall rate than Where2Act and generates more correct interaction directions for the following two reasons: First VRB is trained on human manipulation videos sourced from the real-world domain, directly aligning with the robot's operational environment, whereas Where2Act is trainiend using simulation data. Second, the scale of available human manipulation videos significantly exceeds that of 3D interactable assets, enabling better generalizability.
The reason the recall rate for drawer handles is inferior is that the handles of the drawers used in our work are very small. As a result, they occupy only a small number of pixels in both images and point clouds, making their detection particularly challenging for the two methods. 
{In real-world experiments, we first visualize the predicted action of Where2Act and VRB in the 3D space and manually choose the better one to execute. In over 90\% of scenarios, VRB generates better action than Where2Act.}

\begin{figure}
    \centering
    \includegraphics[width=\linewidth]{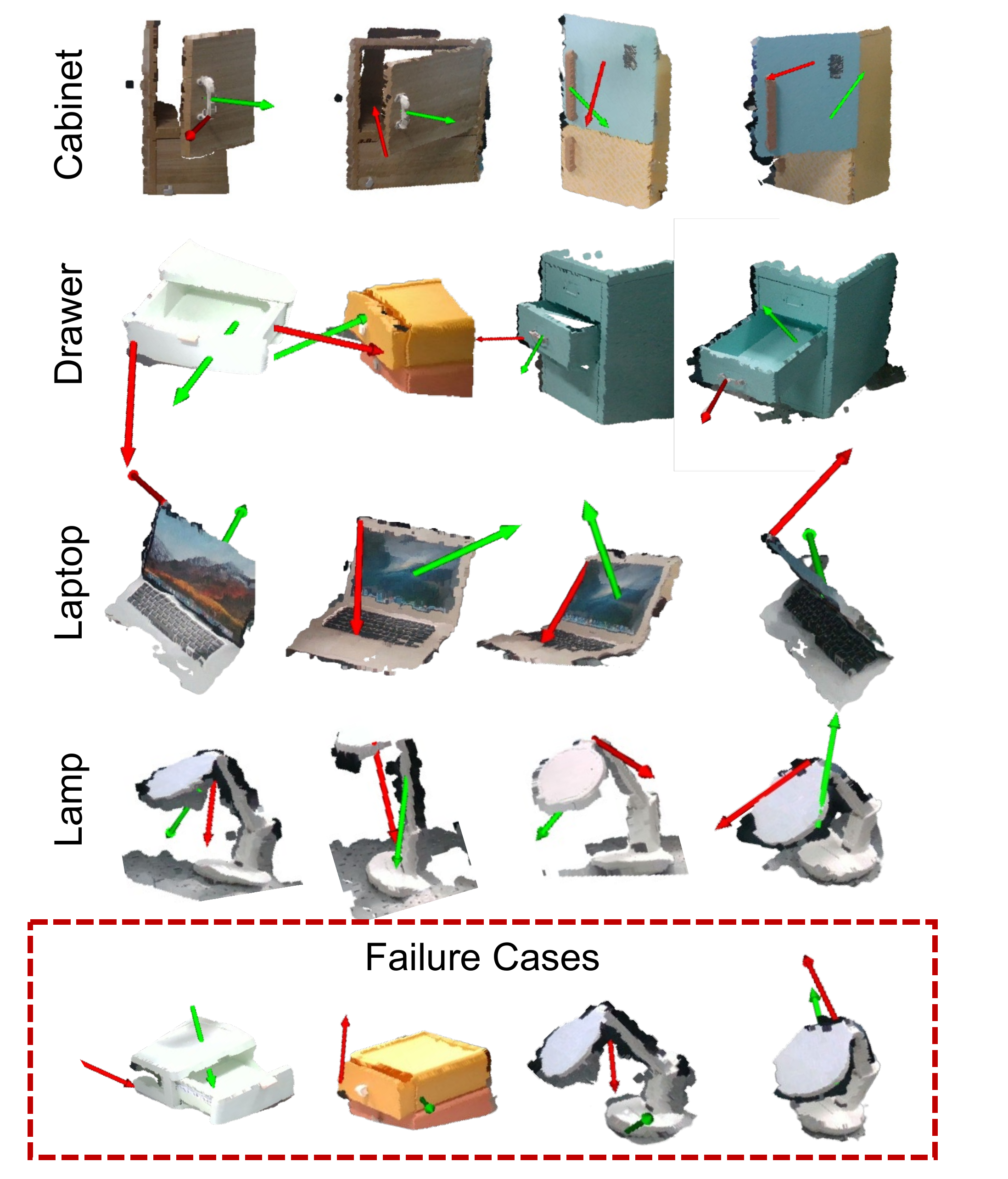}
    \caption{{Visualization of affordance detection results: results of VRB are shown in green arrows, and results of Where2Act are shown in red arrows.}}
    \label{fig:vrb_where2act}
\end{figure}
}
\subsubsection{Pixel Projection}
\begin{figure}[thpb]
      \centering
      \includegraphics[width=8.5cm]{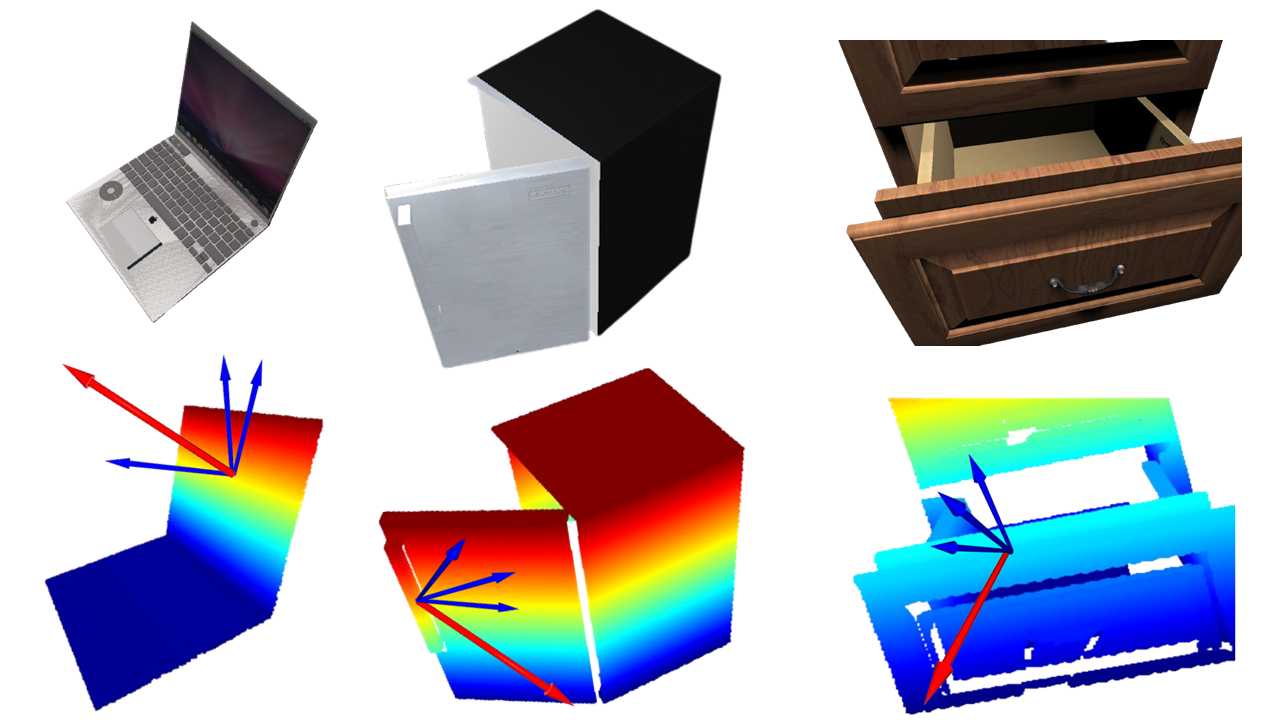}
      \caption{Ablation study on pixel projection. 
      The red vector represents the vector generated by using pixel projection. The three blue vectors are generated only based on 2D affordance.}
      \label{fig:vrb}
\end{figure}

In Section~\ref{sec:method_vrb}, we propose a pixel projection method that leverages both RGB images and depth information of an object to transforms a 2D post-contact vector into a 3D robot trajectory. To evaluate the necessity of the pixel projection approach, we compare it with randomly generated vectors based solely on 2D affordance. Specifically, we select one object per category and generate three random vectors for each object. The results, shown in Fig.~\ref{fig:vrb}, demonstrate that the vector synthesized by pixel transformation is better suited for executing one-step interactions compared to the randomly generated direction vector.

{
\subsection{Objects With Multiple Movable Parts}
In this section, we demonstrate the effectiveness of our method on objects with multiple movable parts. During the interactive perception stage, our method sequentially changes the states of each movable part. Then in the explicit world construction stage, we first separate the generated meshes into movable parts, construct the kinematic structure, and estimate the axis direction and origin of each joint. Fig.~\ref{fig:objects_multiple_parts} illustrates the modeling results of our method applied to two different objects. As shown, our approach accurately reconstructs the explicit models of objects with varying kinematic structures.

\begin{figure}[t]
    \centering
        \subfigure[]{
            \includegraphics[width=3.4cm]{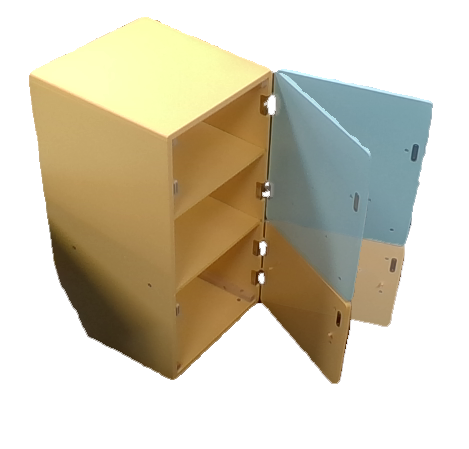}
        }
        \subfigure[]{
            \includegraphics[width=3.4cm]{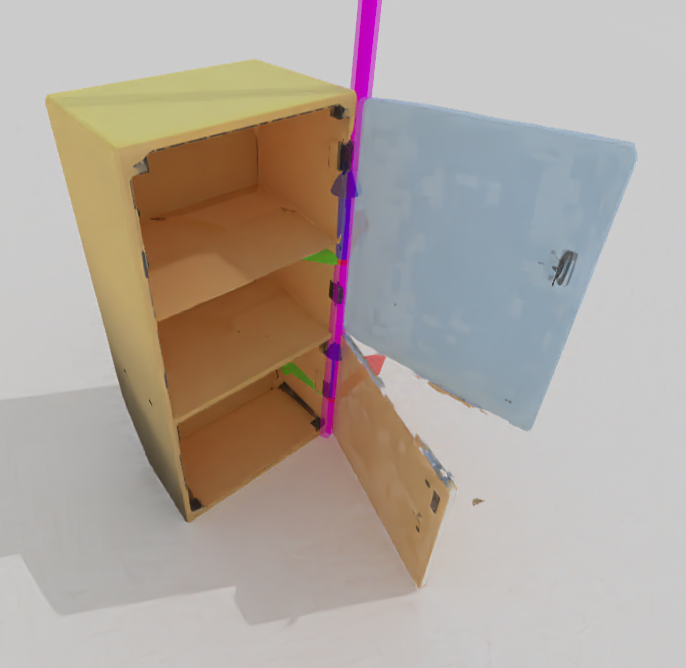}
        }
    \centering
        \subfigure[]{
            \includegraphics[width=3.4cm]{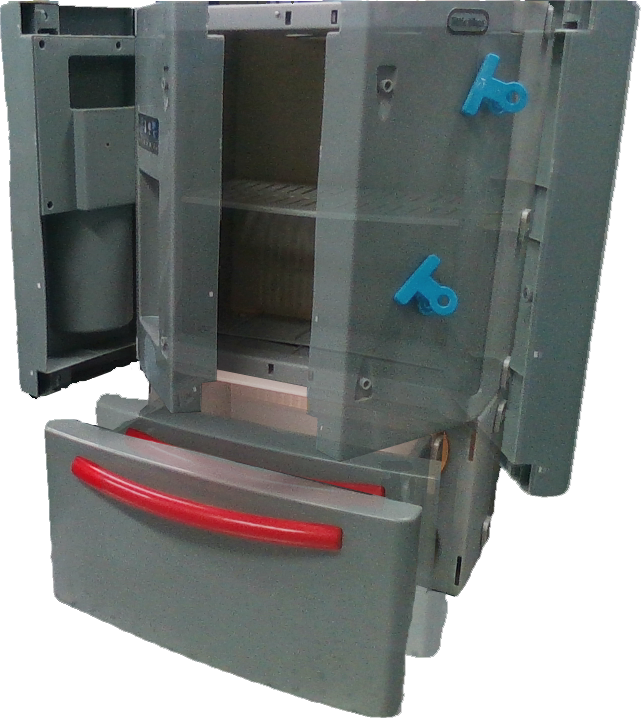}
        }
        \subfigure[]{
            \includegraphics[width=3.4cm]{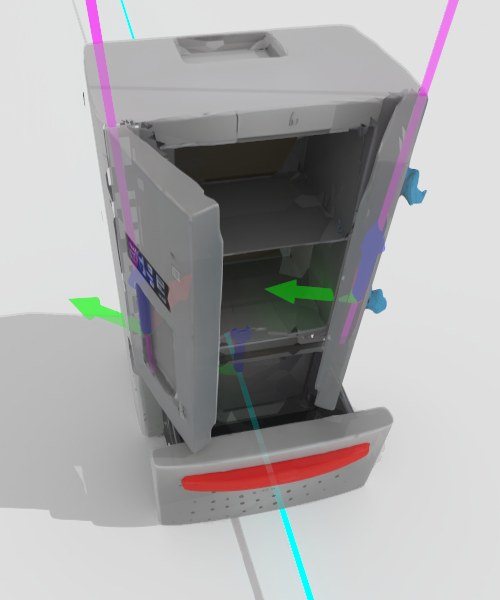}
        }
     \caption{{Multi-frame observations before and after interactions of objects with multiple movable parts and reconstructed articulated models with axes and texture: (a)(b) A cabinet with two revolute joints; (c)(d) A fridge with two revolute joints (shown in pink) and one prismatic joint(shown in blue).}}
     \label{fig:objects_multiple_parts}
\end{figure}
}

\subsection{Advanced Manipulation Skills}

\begin{figure}[t]
    \centering
    \includegraphics[width=\linewidth]{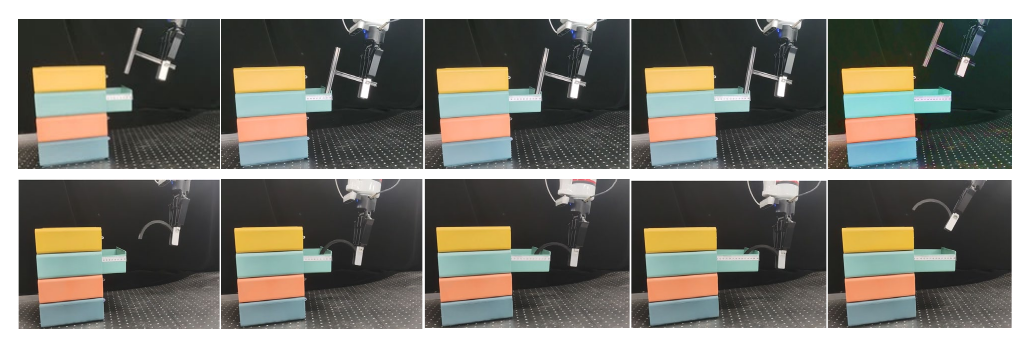 }
    \caption{Open drawer with tools. In real scenarios, the object may be beyond the robot's reach, or the gripper cannot fit into the object's size. Our method can be extended to tool-using cases. As shown in these two sequences, the robot uses a T-shaped tool or a semi-ring to open the small drawer.}
    \label{fig:tool_use}
    
\end{figure}

By utilizing a physics simulation as the explicit world model, our method ensures generalizability to unseen actions. This allows for easy extension to advanced manipulation skills, such as manipulation with tools. As shown in Fig.~\ref{fig:tool_use}, when the drawer is located out of the dexterous range of the robot or the gap between the drawer front and body is too small, the gripper alone cannot open it. In such cases, the robot can employ nearby tools to complete the task.

To demonstrate our method's tool-using capability, we use two different tools for the drawer-opening task. Benefiting from the explicit physics model, we can equip the robot with a tool to interact with the articulated object in the simulation. When using MPC to search for trajectories, we assume the tool is mounted on the robot's end effector. We simply replace the gripper tips with the tool in $r_{dist}$ when computing rewards. Remarkably, our method successfully finds a feasible trajectory with most parameters unchanged. 

{
\section{Analysis and Discussion}
\label{sec:analysis}
In this section, we analyze the reason of manipulation errors and discuss the limitations of our method.

By comparing the manipulation outcomes in simulation and real-world objects, we conclude that the manipulation errors can be attributed to the following factors, in addition to model construction errors:
\begin{enumerate}[label=(\arabic*)]
    \item The dynamic properties of the real articulated objects are complicated. For example, the elastic deformation of laptops, that the laptop partially returns to its original shape once the contact force from the robot is removed, is not modeled in the simulation.
    \item The kinematic structure of a real articulated object is not ideal. For example, there might be gaps in the drawer rails, which transform an ideal prismatic joint into a joint with multiple DoFs.
\end{enumerate}

Our explicit world model construction module infers kinematic structures by analyzing inter-frame differences, which necessitates observable state changes in object parts during the interactive perception stage. Another limitation is that for novel or uncommon objects, existing 3D AIGC techniques may produce meshes that are not perfectly aligned with real-world observations, affecting the accuracy of the resulting digital twins. However, with the rapid advancements in AIGC techniques, we anticipate these alignment challenges will be effectively addressed in the near future.
}
\section{Conclusion}
\label{sec:conclusion}
{
In this work, we present DexSim2Real$^\textbf{2}$, a novel robot learning framework designed for precise, goal-conditioned articulated object manipulation with suction grippers, two-finger grippers and dexterous hands. We first build the explicit world model of the target object in a physics simulator through active interaction and then use MPC to search for a long-horizon manipulation trajectory to achieve the desired manipulation goal. Quantitative evaluation of real object manipulation results verifies the effectiveness of our proposed framework for multiple types of end effectors. 
In this work, we focus on single articulated object manipulation using one fixed robot arm. For future work, we aim to expand the framework to scene-level digital twin construction for long-horizon mobile manipulation. We will also explore more advanced reward function design, such as generating rewards from videos. 

}

\bibliographystyle{IEEEtran}
\bibliography{citation}

\end{document}